\documentclass[sigconf]{acmart}  
\AtBeginDocument{%
  }

\copyrightyear{2025}
\acmYear{2025}
\setcopyright{acmlicensed}
\acmConference[MM '25]{Proceedings of the 33rd ACM International Conference on Multimedia}{October 27--31, 2025}{Dublin, Ireland.}
\acmBooktitle{Proceedings of the 33rd ACM International Conference on Multimedia (MM '25), October 27--31, 2025, Dublin, Ireland}
\acmISBN{979-8-4007-2035-2/2025/10}
\acmDOI{10.1145/3746027.3754777} 

\usepackage{pifont}
\usepackage{listings}
\usepackage{xcolor}
\usepackage{colortbl}
\usepackage{multirow}
\usepackage{subcaption}

\DeclareMathOperator*{\Max}{Max}

\renewcommand{\shortauthors}{Chaoran Feng, et al.}

\fancypagestyle{standardpagestyle}{%
    \fancyhf{} 
    \fancyhead[LE]{MM '25, October 27--31, 2025, Dublin, Ireland} 
    \fancyhead[RO]{Chaoran Feng, et al.} 
}

\begin{document}

\title{E-4DGS: High-Fidelity Dynamic Reconstruction from the Multi-view Event Cameras}

\author{Chaoran Feng}
\authornote{Both authors contributed equally to this research.}
\affiliation{%
  \institution{School of Electronic and
Computer Engineering, Peking University}
  \city{}
  \state{}
  \country{}
}
\email{chaoran.feng@stu.pku.edu.cn}

\author{Zhenyu Tang}
\authornotemark[1]
\affiliation{%
  \institution{School of Electronic and
Computer Engineering, Peking University}
  \city{}
  \state{}
  \country{}
}
\email{zhenyutang@stu.pku.edu.cn}

\author{Wangbo Yu$^*$}
\affiliation{%
  \institution{School of Electronic and
Computer Engineering, Peking University }
  \city{}
  \state{}
  \country{}
}
\email{wbyu@stu.pku.edu.cn}

\author{Yatian Pang}
\affiliation{%
  \institution{National University of Singapore}
  \city{}
  \state{}
  \country{}
}
\email{yatian_pang@u.nus.edu}

\author{Yian Zhao}
\affiliation{%
  \institution{School of Electronic and
Computer Engineering, Peking University}
  \city{}
  \state{}
  \country{}
}
\email{zhaoyian@stu.pku.edu.cn}

\author{Jianbin Zhao}
\affiliation{%
  \institution{School of Future 
Technology, 
  Dalian University of Technology}
  \city{}
  \state{}
  \country{}
}
\email{1518272584@mail.dlut.edu.cn}

\author{Li Yuan}
\authornote{Correspongding Author.}

\affiliation{%
  \institution{School of Electronic and
Computer Engineering, Peking University}
  \city{}
  \state{}
  \country{}
}
\email{yuanli-ece@pku.edu.cn}

\author{Yonghong Tian}
\authornotemark[2]
\affiliation{%
  \institution{School of Electronic and
Computer Engineering, Peking University}
  \city{}
  \state{}
  \country{}
}
\email{yhtian@pku.edu.cn}

\begin{abstract}
Novel view synthesis and 4D reconstruction techniques predominantly rely on RGB cameras, thereby inheriting inherent limitations such as the dependence on adequate lighting, susceptibility to motion blur, and a limited dynamic range.
Event cameras, offering advantages of low power, high temporal resolution and high dynamic range, have brought a new perspective to addressing the scene reconstruction challenges in high-speed motion and low-light scenes.
To this end, we propose \textit{E-4DGS}, the first event-driven dynamic Gaussian Splatting approach, for novel view synthesis from multi-view event streams with fast-moving cameras.
Specifically, we introduce an event-based initialization scheme to ensure stable training and propose event-adaptive slicing splatting for time-aware reconstruction. Additionally, we employ intensity importance pruning to eliminate floating artifacts and enhance 3D consistency, while incorporating an adaptive contrast threshold for more precise optimization.
We design a synthetic multi-view camera setup with six moving event cameras surrounding the object in a 360-degree configuration and provide a benchmark multi-view event stream dataset that captures challenging motion scenarios.
Our approach outperforms both event-only and event-RGB fusion baselines and paves the way for the exploration of multi-view event-based reconstruction as a novel approach for rapid scene capture.
The code and dataset are available on the \href{https://github.com/SuperFCR/E-4DGS}{project page}.
\end{abstract}
\vspace{-5pt}



\begin{teaserfigure}
\vspace{-5pt}
  \includegraphics[width=\textwidth]{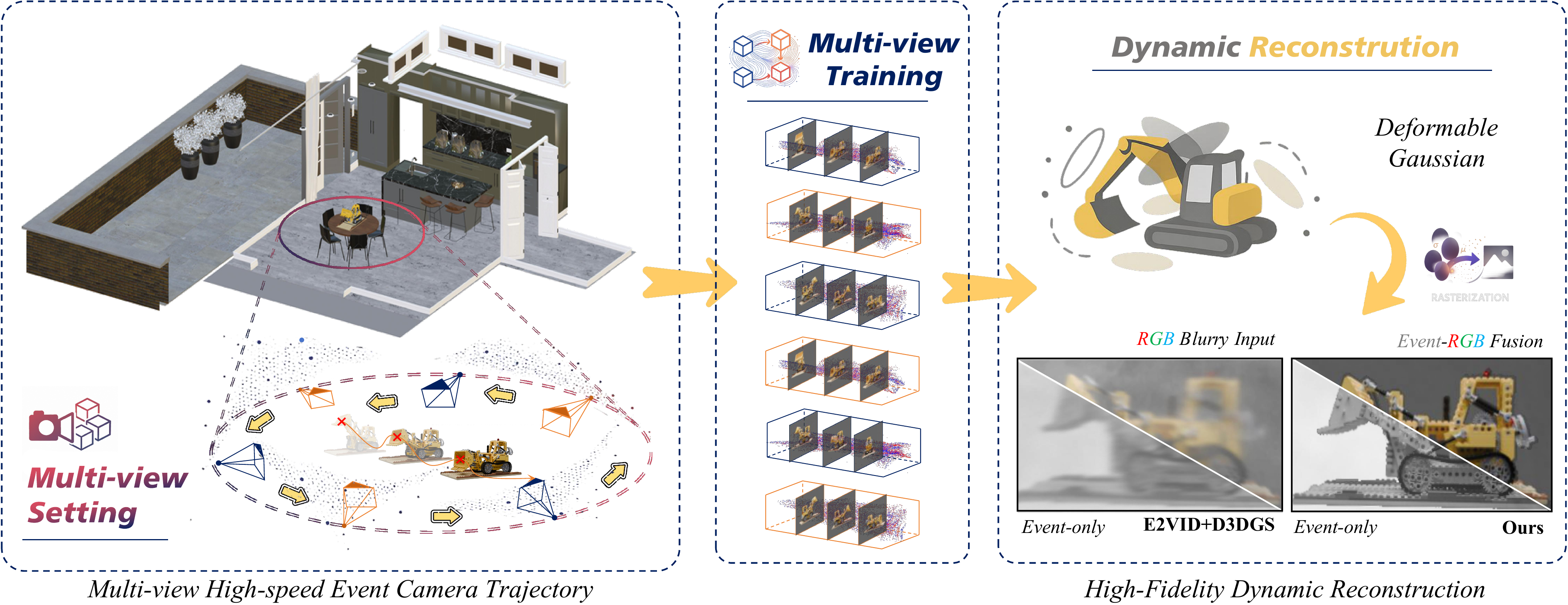}
    \vspace{-2.0em}
    \caption{Our \textbf{E-4DGS} reconstructs temporally consistent and photorealistic dynamic scenes using event streams and sparse RGB frames captured from multi-view moving cameras, effectively handling complex motion and lighting variations.}
  \label{fig:teaser}
\end{teaserfigure}


\renewcommand{\shortauthors}{Chaoran Feng, et al.}  
\maketitle
\section{Introduction}
Novel view synthesis (NVS) and dynamic scene reconstruction are critical for immersive applications such as virtual and augmented reality (VR/AR)~\cite{lee2025dietgsdiffusionpriorevent,yu2025trajectorycrafter}, scene understanding~\cite{chen2023clip2scene-understanding,xue2023ulip-3d-pcl-understanding}, 3D content creation~\cite{deitke2024objaverse,zhang2023repaint123,pang2024dreamdance,tang2024cycle3d,lihao2023freestyleret}, and autonomous driving tasks~\cite{yan2024gs-slam,guo2024cmax-slam,yu2024viewcrafter}. 
While Neural Radiance Fields (NeRF)~\cite{nerf} has recently achieved remarkable success in photorealistic rendering of static scenes, their extension to dynamic scenarios remains challenging—primarily due to substantial training time.
In contrast, 3D Gaussian Splatting (3DGS)~\cite{kerbl3Dgaussians} provides notable advantages in real-time rendering and significantly faster training. 
Yet, existing dynamic extensions of 3DGS struggle to handle scenes with fast motion effectively, primarily due to the inherent limitations of RGB cameras, which, owing to their high latency and limited dynamic range, are prone to motion blur when capturing fast-moving scenes.

Compared to RGB cameras that capture images at fixed intervals, event cameras operate asynchronously by recording brightness changes as event spikes with microsecond-level latency, offering extremely low latency and high dynamic range~\cite{gallego2020survey,wang2025-Event-Cameras-High-mobility-Devices-Survey,eicil_nips_23}
Owing to such advantageous, event cameras have recently been adopted for novel view synthesis and scene reconstruction tasks~\cite{xu2025-event-driven-3d-reconstrution-survey}.
For example, event-driven NeRF methods~\cite{klenk2023e-nerf,rudnev2023eventnerf,hwang2023ev-nerf,feng2025ae-nerf} leverage event accumulation frames and depend on known or estimated camera trajectories to reconstruct NeRF representation. In parallel, event-driven 3DGS approaches~\cite{huang2024inceventgs,wu2024ev-gs,yin2024e-3dgs,han2024event3dgs-nips,li2025gs2e} utilize the sharp structural information provided by event streams to reconstruct 3DGS representation, enabling efficient rendering and training.
However, these methods are primarily designed for static scene reconstruction and are not well-suited for modeling dynamic environments.
In the more challenging task of dynamic scene reconstruction, relying solely on a single event camera inherently limits the ability to capture complete scene dynamics—especially in scenarios involving fast motion, large deformations, or severe occlusions.
Moreover, the coupling between object and camera motion can often lead to mutual cancellation of contrast changes, resulting in neutralized events~\cite{han2024event3dgs-nips,gallego2018unifying-contrast-maximization,ercan2024hypere2vid} that obscure fine-grained geometric details.

Based on the above observation, we aim to investigate the following research question: \textit{How can we efficiently reconstruct a high-fidelity dynamic scenes using multi-view fast-moving event cameras?}
With the captured multi-view event streams, a straightforward approach is to adopt a two-stage pipeline: First, reconstructing intensity frames from the event streams using E2VID~\cite{Rebecq19e2vid,ercan2024hypere2vid} and obtain Gaussian initialization points from COLMAP~\cite{colmap} ; Then, applying an off-the-shelf reconstruction method for futher reconstruction~\cite{yang2024deformable3dgs,duan20244drotorgs,yang20244dgs-iclr2024,yan2024-4dgs-Scale-aware-Residual-Field}.
However, this na\"ive solution compromises the temporal precision and sparsity of event data by converting it into intensity frames, introducing accumulation error and extensive costs, resulting in degraded reconstruction consistency.

To this end, we propose \textit{E-4DGS}, an end-to-end event-based framework for high-fidelity dynamic 3D reconstruction from multi-view event streams. 
To address the initialization challenge under sparse event observations, we introduce an event-specific strategy to generate stable Gaussian primitives without relying on RGB-based Structure from Motion (SfM). 
We further design an event-adaptive slicing mechanism that segments and accumulates event streams for accurate supervision, and propose a multi-view 3D consistency regularization to enhance structural alignment. Additionally, \textit{E-4DGS} supports optional refinement using a few motion-blurred RGB frames.
To our knowledge, this is the first event-only framework enabling view-consistent 3D Gaussian reconstruction in dynamic scenes. For evaluation, we introduce a multi-view synthetic event dataset that serves as a benchmark for dynamic scene reconstruction. 
The dataset encompasses a diverse set of dynamic scenes with simultaneous camera and object motion, ranging from "mild" to "strong".
We compare our method against two-stage baselines that utilize E2VID for intensity reconstruction followed by frame-based methods, trained either with event streams alone or with a combination of RGB videos and event sequences. 
Our approach significantly outperforms all baselines, achieving state-of-the-art results while enabling continuous and temporally coherent reconstruction of dynamic scenes.
These results demonstrate that operating directly on raw event data, especially under challenging conditions with camera motion, yields higher-fidelity dynamic scene reconstruction compared to methods relying on reconstructed RGB frames.
To summarize, the main contributions are as follows:
\vspace{-0.5em}
\begin{itemize}
    \item We present \textit{E-4DGS}, the event-driven approach for reconstructing adynamic 3D Gaussian Splatting representation from multi-view event streams.
    
    \item We introduce an event-based initialization scheme for stable training, propose event-adaptive slicing splatting and adaptive event threshold for supervision, and design intensity importance pruning to enhance 3D consistency.
    
    \item We construct a multi-view synthetic dataset with moving cameras for 4D reconstruction from event streams. Our method achieves state-of-the-art performance, and we will release our work to support future research.
\end{itemize}

\section{Related Work}

\subsection{Dynamic Reconstruction from RGB Frames}
Modeling dynamic scenes from moving RGB cameras alone is still a challenging open task in computer vision. 
A widely used approach to this problem is to learn coordinate-based neural scene representations allowing rendering novel views and representing dynamic scenes.
Previous works such as neural radiance field (NeRF) and its variants D-NeRF~\cite{pumarola2021d-nerf} and more~\cite{park2021hypernerf,park2021nerfies,cao2023hexplane,yan2023nerf-ds,lin2024dynamicnerf-review} used implicit neural representations in combination with volume rendering.
They are based on Multi-Layer Perceptrons (MLPs), which are relatively compact and require minimal storage space once trained. 
However, they are expensive to optimize and lead to slow training and evaluation which limits its expansion on the real-time rendering and real-world applications.
The recently emerging 3DGS~\cite{kerbl3Dgaussians} and its variants~\cite{mip-splatting-cvpr24,tang2025neuralgs,chen2024usp-gs,guo2024spikegs} have reshaped the landscape of dynamic radiance fields due to its efficiency and flexibility.
The pioneering work Deformable3DGS (D3DGS)~\cite{yang2024deformable3dgs} enhances dynamic Gaussian representations with a tiny deformable field for tracking the motion of Gaussian points.
Similarly, other methods~\cite{wang2024shape-of-motion,he2024s4d,wu2025swift4dgs,shan2025deformablegs-efficient-icra25,yang20244dgs-iclr2024,pang2024next,wan2024template-free-4dgs-nips2024,lu2025dn-4dgs} models Gaussian motion using point-tracking functions for stable point moving.
Our approach adopts D3DGS as the dynamic representation due to its simple and efficient structure, and then presents its application to the supervision from event streams. It inherits thereby the advantages of event streams and 3DGS for dynamic view synthesis.

\label{sec:preliminaries}
\begin{figure*}[t]
    \centering
    \includegraphics[width=1.0\linewidth,trim={0cm 0cm 0cm 0cm},clip]{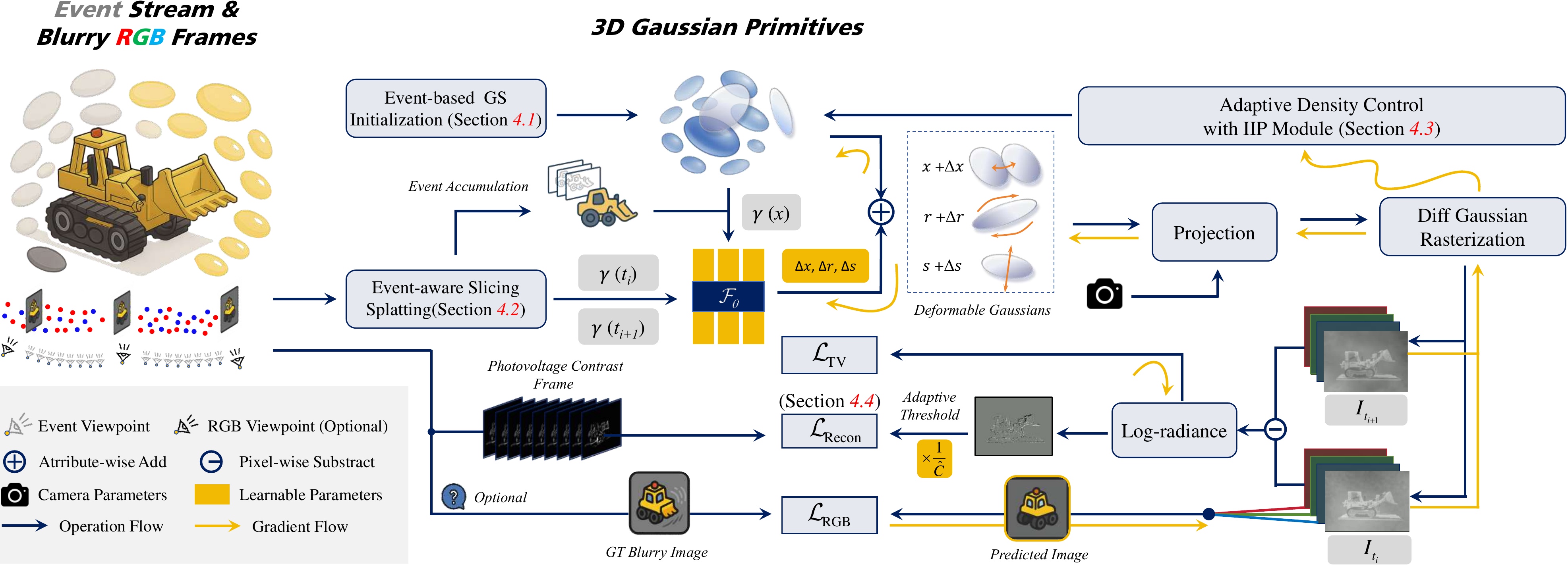}
    \vspace{-1.5em}

    \caption{
        The overview of E-4DGS. Our framework establishes temporal-coherent 4D representations through a cascaded processing of event streams: 
        The event-driven initialization constructs spatio-temporal gaussians via center-focus density fields, followed by differentiable feature distillation where adaptive slicing operators disentangle high-frequency part for GS optimization. 
        Cross-view consistency is then imposed through deformable Gaussian reprojection coupled with saliency pruning, while multi-modal alignment ultimately achieves photometric fidelity via kernel-attentive RGB-event synchronization.
    }
    \vspace{-1.5em}
    \label{fig:pipeline}
\end{figure*}

\subsection{Dynamic Reconstruction from Event Data}
Event cameras have been widely used to reconstruct dynamic scenes from non-blurry RGB Frames of fast motion.
Previous works, including model-based methods~\cite{pan2019edi, rudnev2023eventnerf} and learning-based methods~\cite{Rebecq19e2vid, tulyakov2021event-time-lens, he2022event-timereplayer}, process event and RGB frames with 2D priors but lack 3D consistency.
Other event-based methods address tasks such as detection, tracking, and image/3D reconstruction, including lip reading~\cite{tan2022event-spatio-temporal-lip-cvpr22, savran2018-event-lip-energy}, object tracking~\cite{wu2024event-leod-detection-cvpr24, zhao2023event-object-detection-icra23}, and pose estimation~\cite{zou2021event-3dhpe-cvpr21, goyal2023event-hpe-moveenet-cvpr23,li2025gs2e}. 
However, these methods still do not incorporate 3D priors to reconstruct scene appearance and are not applicable to represent 3D scenes, which is our goal of proposed \textit{E-4DGS}.

For static scene reconstruction, recent event-based methods~\cite{xiong2024event3dgs, evagaussians, yin2024e-3dgs, evdeblurnerf, tang2024lse-nerf, low2023robust, wang2024evggs, zhang2024elite-ev-gs, wu2024-sweepevgs, huang2024inceventgs, qi2023e2nerf,evagaussians,lihao2025decoupled,yuanshenghai2024magictime} have achieved high-fidelity 3D reconstruction and NVS tasks using supervision from event pixels or event accumulation. 
These methods primarily rely on consistent event sequences from a single mono-event camera. 
However, extending static scene representations to dynamic scenes with event streams is a challenging task, as the movement of objects and the simultaneous motion of the event camera can introduce ambiguity in the events. 
Different from only a single mono-camera setting, our proposed E-4DGS reconstructs the dynamic scene with the multi-view camera setting, providing more multi-view consistency details.

Recently, a growing trend is the use of dynamic neural radiance fields (DNeRF) or Dynamic 3DGS (4DGS) for dynamic scene representation and novel view synthesis.
DE-NeRF~\cite{ma2023de-nerf} and EBGS~\cite{xu2024event-boosted} reconstruct dynamic scenes using monocular event streams and RGB frames from a moving camera, modeling deformations in a canonical space. The former is based on DNeRF, while the latter relies on 4DGS. 
EvDNeRF~\cite{bhattacharya2024evdnerf}, which is based on canonical volumes, and DynEventNeRF~\cite{rudnev2024dynamic-eventnerf}, which uses temporally-conditioned MLP-based NeRF, both utilize multi-view event streams to reconstruct dynamic scenes.
However, the former does not model appearance, and the latter is trained slowly due to volume rendering. 
In contrast, our proposed E-4DGS achieves higher-quality reconstruction by accurately capturing complex geometries and lighting effects than NeRF-based models, while also offering fast training and inference speeds for real-time, real-world applications.
\section{Preliminaries}

\subsection{Deformable 3D Gaussian Splatting}
\label{sec:preliminaries:dynamic3dgs}

Deformable3DGS~\cite{yang2024deformable3dgs} offers an explicit method for representing a 4D dynamic scene $\mathbb{G}$ with the canonical space and th deformable space based on 3D Gaussian Splatting~\cite{kerbl3Dgaussians}.
In the canonical space, these 3D Gaussian points have the following parameters: mean point $\mu$, covariance matrix $\Sigma$, opacity $\sigma$, and color $\bf{c}$ and a 3D Gaussian point $G(x)\in \mathbb{G}$ is defined as follows:
\begin{equation}
    \label{eq:preliminary:3dgs}
    G(x) = e^{-\frac{1}{2}(x-\mu)^T \Sigma^{-1} (x-\mu)}
\end{equation}
where, $\Sigma$ is divided into two learnable components: the quaternion $r$ represents rotation, and the 3D-vector $s$ represents scaling.
then, the color of each pixel can be calculated using the following formula:
\begin{equation}
    \label{eq:preliminary:C_alpha}
    C(x) = \sum_{i \in \mathcal{N}(x)} c_i \alpha_i(x) \prod_{j=1}^{i-1} \left( 1 - \alpha_j(x) \right),
\end{equation}
where $\alpha_i(x) = \sigma_i \exp \left( -\frac{1}{2}(x - \mu_i^{2D})^T \Sigma^{-1}(x - \mu_i^{2D}) \right)$, and $N$ is the number of Gaussian points that intersect with the pixel $x$.

In the deformable space, Deformable3DGS employ a compact MLP layer to represent motion of Gaussian points.
Given timestamp $t$ and center position $x$ of 3D Gaussians as
inputs, the deformation MLP produces offsets, which subsequently transform the canonical 3D Gaussians to the deformed space:
\begin{equation}
    \label{eq:preliminary:deformable_space}
    (\Delta x, \Delta  r, \Delta  s) = \mathcal{F}_{\theta} (\gamma(\text{sg}(x))), \gamma(t))
\end{equation}
where $sg(\cdot)$ indicates a stop-gradient operation, $\gamma$ denotes the positional encoding as defined in~\cite{yang2024deformable3dgs}. 
Therefore, a dynamic Gaussian point can be represented as $G(x+\Delta{x},r+\Delta{r},s+\Delta{s})$ at timestamp $t$.

\subsection{Event Generation Model}
\label{sec:preliminaries:egm}
A single event is represented as $e_k = (x_k, y_k, p_k, t_k)$ in the event streams $\mathcal{E}$,
denoting a brightness change registered by an event sensor at timestamp $t_k$, pixel location $\mathbf{u_k}=(x_k, y_k)$ in the event camera frame with polarity $p_k \in \{-1,+1\}$.
The change between adjacent timestamps can be calculated from intensity images $I$.
\begin{align}
      L(\mathbf{u}_k, t_k) - L(\mathbf{u}_k, t_{k-1})&=  \sum_{t_{k-1} < t \leq t_{k}} p_t{C^{p_t}} \overset{\text{def}}= {\Delta}E_{\mathbf{u}_k}(t_{k-1}, t_{k}), \label{eq:preliminaries:delta_L} \\
    \text{where} \quad L &= \log (I). \label{eq:preliminaries:log_I}
\end{align}
Here, the thresholds $C^{p}\in\{C^{-1}, C^{+1}\}$ define boundaries for classifying the event as positive or negative, with the polarity of an event indicating a positive or negative change in logarithmic illumination.

Therefore, given a supervisory event stream $\mathcal{E}$, we can supervise our proposed \textit{E-4DGS} by comparing the predicted brightness change ${\Delta}\hat{E}(t_{k-1}, t_{k})$ and the ground truth ${\Delta}E(t_{k-1}, t_{k})$ by Equation~(\ref{eq:preliminaries:delta_L}) over all image pixels. 
In general, we substitute intensity frames $\hat{I}_t$ with the rendered results $\hat{C}_t$ and can utilize photo-realistic loss~\cite{yang2024deformable3dgs} between the predicted intensity frames and the ground-truth event of event-based single integral (ESI)~\cite{rudnev2023eventnerf}:
\begin{equation}
    \label{eq:preliminaries:event_gs_loss}
    \mathcal{L}_{gs}=\sum_{{\mathbf{u}_k}\in\hat{I}}({\lambda{\mathcal{L}_1({\Delta}\hat{E}_{\mathbf{u}_k},{\Delta}E_{\mathbf{u}_k})}+(1-\lambda)\mathcal{L}_{D-SSIM}({\Delta}\hat{E}_{\mathbf{u}_k},{\Delta}E_{\mathbf{u}_k})})
\end{equation}

\section{Method}
\label{sec:3:method}
We propose \textbf{E-4DGS}, a method for high-fidelity dynamic scene reconstruction using sparse event camera streams. 
Given multi-view event data capturing a dynamic scene, E-4DGS reconstructs a 4D model that allows novel view generation at arbitrary times. 
To address the challenges posed by the sparse nature of event data and the dynamic characteristics of the scene, we introduce an event-based initialization strategy (Section\textcolor{red}{~\ref{sec:3:method:event-based-initialization}}), an event-aware slicing splatting technique to preserve geometric details (Section\textcolor{red}{~\ref{sec:3:method:event-adaptive-slicing-splatting}}), and multi-view 3D consistency regularization for improved scene fidelity (Section\textcolor{red}{~\ref{sec:3:method:intensity-importance-prunning}}). 
Additionally, we utilize adaptive event supervision and color recovery to enhance the reconstruction quality (Section\textcolor{red}{~\ref{sec:3:method:event-supervision-optimization}}). 
The overview of our method is illustrated in Figure~\ref{fig:pipeline}.




\subsection{Event-based Initialization}
\label{sec:3:method:event-based-initialization}
The Gaussian primitives are initialized using a point cloud derived from Structure-from-Motion (SfM)~\cite{micusik2006sfm} with RGB frames in the vanilla 3DGS. 
However, their performance is hindered by inaccurate dynamic Gaussian initialization due to view inconsistencies caused by object motion.
Furthermore, applying SfM to extract Gaussian points from event sequences is more challenging than using RGB frames with COLMAP~\cite{colmap}, due to the sparse nature of event streams. 
Some methods~\cite{huang2024inceventgs, wu2024ev-gs, han2024event3dgs-nips, xiong2024event3dgs} randomly initialize Gaussians within a fixed cube without considering unbounded scenes. 
Other methods perform better than random initialization but are more complex. Elite-3DGS~\cite{zhang2024elite-evgs} employs a two-stage approach with E2VID~\cite{Rebecq19e2vid} to convert events into images, followed by SfM for point cloud initialization, while E-3DGS~\cite{yin2024e-3dgs} uses exposure enhancement~\cite{snavely2006photo-tourism} method before obtaining the SfM points.


Thus, we adopt an event-specific strategy for Gaussian point initialization, balancing performance and efficiency. 
Specifically,
1) For object scenes, we initialize the point cloud with 100,000 points in a fixed cube, consistent with original 3DGS settings;
2) For medium or large scenes, we employ a dense-to-sparse radiative sphere initialization, mimicking realistic distribution where point density is highest at the center and decreases toward the boundaries. We set sphere's radius to $r = 10.0$ with 200,000 initial points.

We also experimented with initializing the Gaussian primitives using random pointcloud and E2VID+COLMAP, and further details are provided in the supplementary materials. 
While our approach yielded a slight performance drop than the E2VID+COLMAP's performance, the latter requires more computational complexity.
%

\subsection{Event-adaptive Slicing Splatting}
\label{sec:3:method:event-adaptive-slicing-splatting}
In event-based scene reconstruction pipelines, the slicing strategy for the event stream significantly influences reconstruction quality. 
As the duration of the event time window $(t_{i},t_{i+1})$ increases, 
the predicted events become a discretized, aliased representation of the continuous brightness variations in the scene.
\begin{figure}[t]
    \centering
    \includegraphics[width=1.0\linewidth,trim={0cm 0cm 0cm 0cm},clip]{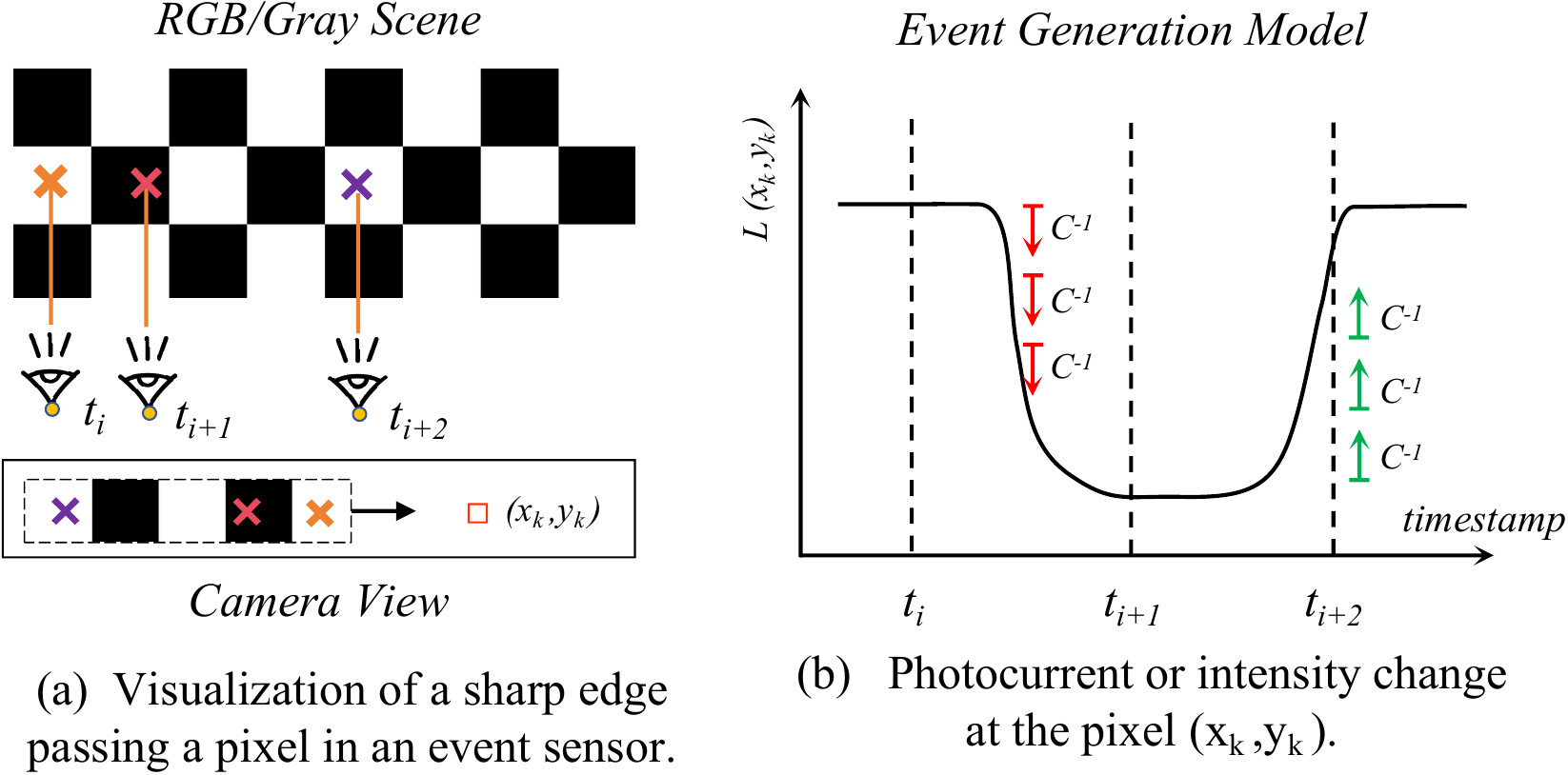}
    \vspace{-1.5em}
    \caption{
        Demonstration of how time window discretizations can influence the count of events between timestep pairs. The time window $(t_{i}, t_{i+1})$ produces two negative events, whereas $(t_{i}, t_{i+2})$ results in event neutralization.
    }
    \vspace{-1.5em}
    \label{fig:time_window_visualization}
\end{figure}

For instance, Figure~\ref{fig:time_window_visualization} illustrates that measurements recorded by the event sensor between timestamps $(t_i, t_{i+1})$ produce three negative events at the selected pixel, whereas measurements over the interval $(t_i, t_{i+2})$ yield no events. 
This effect is particularly notable in our pipeline, as the process of accumulating polarity inherently neutralizes events. 
Moreover, existing works~\cite{xiong2024event3dgs, rudnev2023eventnerf} have demonstrated that using consistently short windows impedes the propagation of high-level illumination, while consistently long windows often result in a loss of local detail.
While they randomly sampled the length of event timestamp window, a drawback is that it does not take into account the camera speed or event rate, causing the sampled windows to contain either too many or too few events. 
Additionally, Hu et al.~\cite{vid2e} and Han et al.~\cite{han2024physical-BECS-simulator} revealed that regions with uniform and smooth intensities typically do not trigger any events, leading to spatial sparsity in the event streams used as supervisory signals.

Based on the aforementioned observations, we propose an event-adaptive slicing strategy to address this issue. 
Specifically, during the training of our \textit{E-4DGS}, we deliberately vary the time window of batched events and incorporate event noise during the event accumulation process. Notably, these modifications lead to an improved generation of finely-sliced events at test time. 
The detailed process of event-adaptive slicing are as follow:

1) \textit{Event Accumulation Range Setting:} For each timestamp, we randomly sample and slice a target number of events streams within the event count range $[N_{min}, N_{max}]$. 

2) \textit{Event Accumulation Jitter:} During our sampling process, we add Gaussian noise to pixels that do not record any events within the whole event timestamp window. This augmentation enhances gradient optimization in smooth regions and increases overall robustness of the pipeline against noisy events. It serves as event sampling in \cite{ma2023de-nerf}, and the whole process is defined as:
\newcommand{\triggered}{\mathbb{1}_{\text{trig}}}  

\begin{equation}
    \label{eq:event_accumulation}
    {\Delta}E_{\textbf{u}}(t_{start},t_{end}) = \begin{cases}
\displaystyle \int_{t_s}^{t_e} p_{\tau }{C^{p_{\tau}}}, d\tau 
& \text{if } \triggered \neq 0, \\[8pt]
\Delta \cdot \mathcal{N}\bigl(0, \sigma_{\text{noise}}^2\bigr) 
& \text{if } \triggered = 0.
\end{cases}
\end{equation}

where, $\Delta{E(\cdot)}$ denotes the event frame accumulated from all event polarities triggered at pixel coordinate $\mathbf{u}$ within the current event time window. 
$\triggered$ denotes the spiking of the events.
$t_{\text{start}}$, $t_{\text{end}}$, and $\Delta t = t_{\text{end}} - t_{\text{start}}$ represent the start timestamp, end timestamp, and the time interval of the event time window, respectively.

This strategy not only guarantees a diverse range of event window lengths, but also curtails the loss of fine details that can occur due to neutralization. 
Moreover, it helps preserve critical geomerty details, thereby enhancing the overall fidelity of the reconstruction.

\subsection{Intensity Importance Pruning}
\label{sec:3:method:intensity-importance-prunning}
In the vanilla Gaussian Splatting pipeline, the opacity of all Gaussian points is gradually reduced, and points with low transparency are pruned during the Gaussian pruning stage. 
However, this method is unsuitable for our event-based approach, as it results in excessive coupling between the canonical and deformation fields and simultaneous camera and object motion, further exacerbating the issue.
Therefore, we eliminate the reset opacity operation same as in~\cite{fan2024instantsplat}. 
and drawing inspiration from LightGaussians~\cite{lightgaussian}, which emphasizes a compact representation of static scenes by pruning redundant Gaussians based on spatial attributes such as transparency and volume, we adopt a specialized strategy, \textit{Intensity Importance Pruning} (IIP), to remove floaters across both the canonical and deformable spaces. 
With this strategy, the importance of each Gaussian point is computed for each training viewpoint at every timestamp. 
Gaussian primitives with an importance score below a fixed threshold are then pruned, effectively mitigating the floater issue and enhance the 3D consistency from multi-view event streams.

\begin{figure}[t]
    \centering
    \includegraphics[width=1.0\linewidth,trim={0cm 0cm 0cm 0cm},clip]{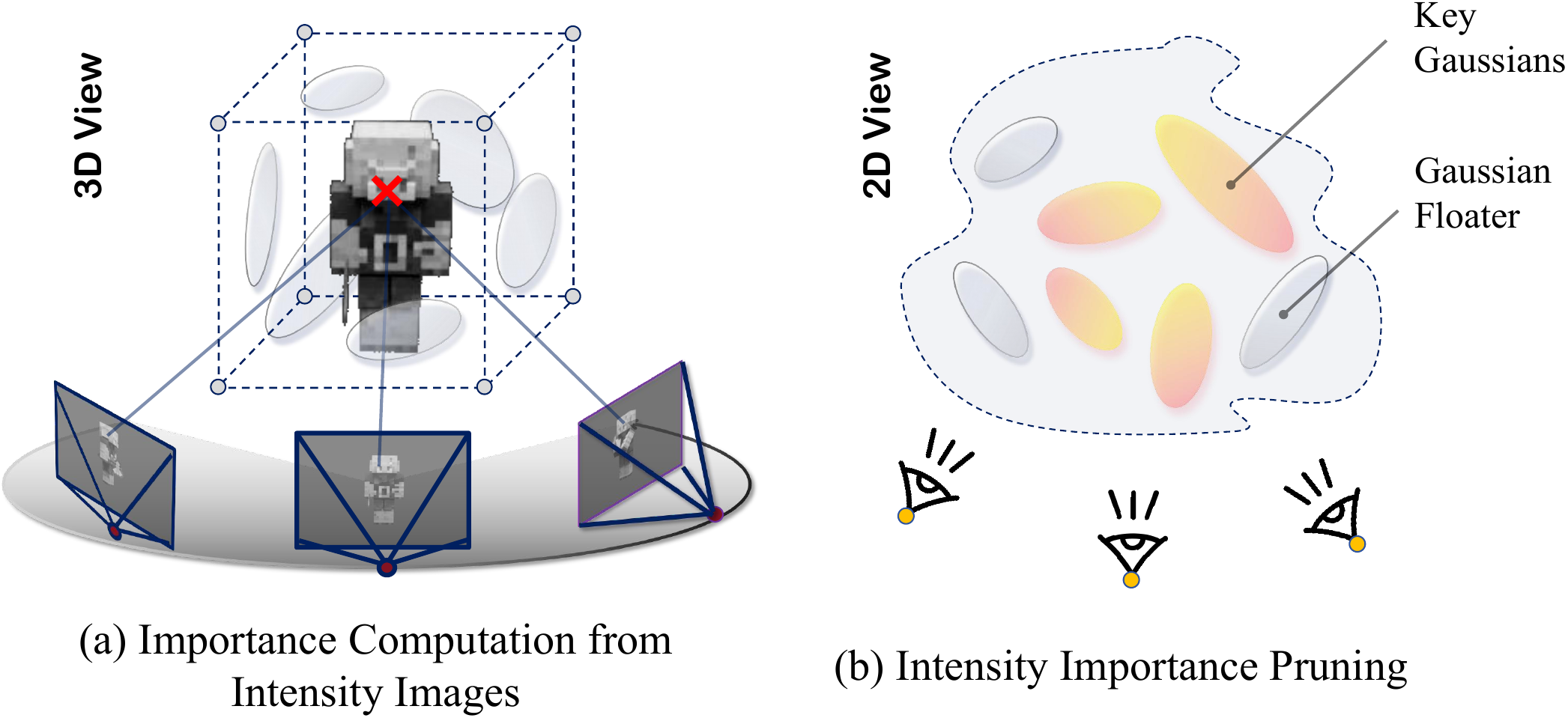}
    \vspace{-2.0em}
    \caption{
        The process of the intensity importance pruning.
    }
    \vspace{-10pt}
    \label{fig:intensity_importance_pruning}
\end{figure}

Specifically, for a Gaussian point $ g_{i} \in \mathbb{G} $, the Gaussian importance $ w_i $ over the images $\mathcal{I}$ of all training views and timestamps $\mathcal{T}$, is defined as follows:
\begin{equation}
    \label{eq:intensity_importance_pruning}
    w_i = \Max_{\mathbf{x} \in I,\, t \in \mathcal{T}}
    \left( \alpha_i\bigl(\mathbf{x} \mid t\bigr) \prod_{j=1}^{i-1} \Bigl( 1 - \alpha_j\bigl(\mathbf{x} \mid t\bigr) \Bigr) \right).
\end{equation}

Here, $ I \in \mathcal{I} $ denotes the intensity image. We prune Gaussian points whose importance scores satisfy $ w_i < 0.015 $, following the approach~\cite{lightgaussian}.
As shown in Figure~\ref{fig:intensity_importance_pruning}, our method effectively removes floating artifacts absent from the training views. 
In addition, we perform Gaussian cloning and splitting following the 3DGS protocol, ensuring that child Gaussian points inherit the dynamic characteristics of their parent Gaussian points.

\subsection{Event Supervision and Optimization}
\label{sec:3:method:event-supervision-optimization}

\textbf{Adaptive Event Supervision.}  
According to previous work~\cite{lin2022dvs-voltmeter}, the ground truth of the event contrast inherently contains some errors. Additionally, in the real scenes captured by an event sensor, the event constrast threshold $C^{p}$ varies due to the environment disturbance, which can make Equation.~\ref{eq:preliminaries:log_I} impractical to use in a real-world setup.
Thus, if we directly apply the photometric loss with Equation.~\ref{eq:preliminaries:event_gs_loss} to compare the rendered intensity frames with those derived from event data, the inherent discrepancies will be strictly penalized during optimization, which may in fact degrade the overall reconstruction quality.
To bridge the gap between synthetic and real event data, we introduce learnable threshold parameters $\hat{C}$ and compute the rendered intensity frame as follows:
\begin{equation}
    \label{eq:adaptive_threshold}
    {\Delta}\hat{E}_{\mathbf{u}_k}(t_{k-1},t_{k}) = \hat{L}(\mathbf{u}_k, t_k) - \hat{L}(\mathbf{u}_k, t_{k-1})\overset{\text{def}}\simeq  \sum_{t_{k-1} < t \leq t_{k}} p_t{\hat{C}},
\end{equation}
Here, we can simpfy this process as follows:

\begin{align}
    N_{gt}(\cdot)_{t_{k-1}}^{t_{{k}}} &= \frac{1}{C} \left( (V(\cdot, t_2) - V(\cdot, t_1)) \right), \label{eq:event_contrast_gt} \\
    N_{pred}(\cdot)_{t_{k-1}}^{t_{{k}}} &= \frac{1}{\hat{C}} \left( (\hat{L}(\cdot, t_{k}) - \hat{L}(\cdot, t_{k-1})) \right),\label{eq:event_contrast_pred} \\
    \mathcal{L}_{Recon} &=  \frac{1}{H \times W} \sum_{{\mathbf{u}} \in \hat{L}} \sqrt{ (N_{gt}(\mathbf{u}) - N_{pred}(\mathbf{u}))^2 + \epsilon^2 }
\end{align} 
Here, $V(\mathbf{u}, t)$ denotes the photovoltage in event pixel $\mathbf{u}$ at timestamp $t$, and $\epsilon$ is a small constant added for numerical stability. $N_{gt}(\mathbf{u})$ and $N_{pred}(\mathbf{u})$ represent the g.t. and predicted event count maps over $(t_{k-1}, t_k]$y. 
The overall event supervision loss is given by:
\begin{equation}
    \label{eq:total_loss}
    \mathcal{L}_{\text{Event}} = \lambda_{\text{Recon}} \mathcal{L}_{\text{Recon}} + \lambda_{\text{TV}} \mathcal{L}_{\text{TV}},
\end{equation}

where $\mathcal{L}_{\text{TV}}$ is a total variation regularization term encouraging spatial smoothness, and $\lambda_{\text{Recon}}$, $\lambda_{\text{TV}}$ are weighting factors balancing the contributions of each component.

\textbf{Combined Gain and Offset Correction.}
Since event cameras only capture logarithmic intensity differences rather than absolute log-intensity values, the predicted log-intensity $\hat{L}$ from our 4DGS method is determined only up to an additive offset for each color channel.
Moreover, there is a scale ambiguity in the reconstructed color balance and illumination of the scene, when only the event contrast threshold is known.
Thus, it's necessary to correct and align the color value for every color channel like previous works~\cite{rudnev2023eventnerf,low2023robust,zahid2025e3dgs-3dv}, using the correction formula as follow:
\begin{equation}
    \label{eq:color_correction}
     \hat{L}(\mathbf{u}_k, t_k) \overset{\text{def}}= g_c \cdot \hat{L}(\mathbf{u}_k, t_k) + \Delta{c},
\end{equation}
where, $g_c$ and $\Delta{c}$ are the color correction parameters, and derived via ordinary least squares~\cite{low2023robust} with the ground-truth log-intensity $L(\mathbf{u}_k, t_k)$ as defined in Section~\ref{sec:preliminaries:egm}.
Notably, the images captured by a separate standard camera are affected by saturation in real-world scenes due to its limited dynamic range, and they are not raw recordings but have undergone lossy in-camera image processing.
Moreover, the contrast threshold of real event cameras varies spatially across the image plane and temporally over time~\cite{vid2e}, making accurate color correction challenging and potentially leading to misalignment in the synthesized views of real scenes.

\section{Experiments}
\label{sec:experiments}

\subsection{Experimental Setting}
\noindent \subsubsection{Implementation Details.} 
(1) Training Assumption:
To reconstruct dynamic scenes using Gaussian Splatting~\cite{kerbl3Dgaussians} from high-speed, multi-view event cameras, we assume that our method leverages accurate camera intrinsics and high-quality, frequency-consistent extrinsics to enable precise interpolation at arbitrary timestamps. Specifically, we apply linear interpolation for camera positions and spherical linear interpolation (SLERP) for camera rotations
For the synthetic event dataset, we adopt the original contrast thresholds $C^{+1}$ and $C^{-1}$ from the v2e simulation settings~\cite{vid2e}. 
In the real-world autonomous driving dataset, we initialize the contrast thresholds using expected values of the event camera settings. 
This prior assumption provides a stable starting point, leading to more consistent training and improved 3D reconstruction performance.

(2) Training Details:
We implemented E-4DGS based on the official code of Deformable3DGS~\cite{yang2024deformable3dgs}, Gaussianflow~\cite{gaussianflow}, E-NeRF~\cite{klenk2023e-nerf} and Event3DGS~\cite{han2024event3dgs-nips,xiong2024event3dgs} with Pytorch and conduct all experiments on a single NVIDIA RTX 4090 GPU.
During training, we render at a resolution of $346 \times 260$ for the synthetic dataset and retain the original resolution $640 \times 480$  for real-scene data. 
Events are accumulated into frames using our adaptive slicing strategy (Section\textcolor{red}{~\ref{sec:3:method:event-adaptive-slicing-splatting}}), where the number of events per temporal window is randomly sampled from a predefined range to introduce temporal diversity and enhance robustness. 
Specifically, we set $[N_{\text{min}}, N_{\text{max}}] = [5\times10^{3}, 10^{4}]$ for object-level scenes and $[10^{5},10^{6}]$ for indoor or large-scale scenes. 
Additionally, Gaussian noise ($\sigma_{\text{noise}} = 0.02$) is injected into event-void pixels during accumulation to improve optimization in textureless regions.
Each scene is trained for 50{,}000 iterations using the Adam optimizer. 
The overall loss consists of an event-based supervision loss, a total variation regularization term and a RGB reconstruction loss (opt. ), weighted by $\lambda_{\text{Recon}} = 1.0$, and $\lambda_{\text{TV}} = 0.005$,  $\lambda_{\text{RGB}} = 1.0$, respectively. 
The stabilization constant $\epsilon$ in $\mathcal{L}_{\text{Recon}}$ is set to 0.001. 
The learnable event contrast threshold $\hat{C}$ is initialized to 0.15 for synthetic scenes and 0.2 for real-scene scenes, and is jointly optimized during training. 
To prevent interference with dynamic scene modeling, opacity reset is disabled like in~\cite{fan2024instantsplat} and color correction is applied only at inference time in all scenes.

\noindent \subsubsection{Evaluation Metrics.} 
For synthetic and real-scene datasets, we employ the Peak Signal-to-Noise Ratio (PSNR), Structural Similarity Index Measure (SSIM) , and VGG-based Learned Perceptual Image Patch Similarity (LPIPS) to evaluate the similarity between rendered novel views and ground-truth novel views. 

\noindent \subsubsection{Baselines.}
At the time of writing, the event-based dynamic reconstruction methods Dynamic EventNeRF~\cite{rudnev2024dynamic-eventnerf} and EBGS~\cite{xu2024event-boosted} have not been publicly released. 
Although EvDNeRF~\cite{bhattacharya2024evdnerf} is open-sourced, it focuses solely on modeling geometric edges rather than performing holistic scene reconstruction.
Consequently, we compare our proposed method against RGB-based baselines that do not utilize event data and are trained either on blurry RGB recordings or on RGB videos reconstructed from events using E2VID~\cite{Rebecq19e2vid}. We choose Deformable3DGS~\cite{yang2024deformable3dgs} and Deblur4DGS~\cite{Deblur4DGS} as the RGB-based baseline with blurry RGB inputs or event-integral inputs.

\begin{figure*}[!t]
    \centering
    \includegraphics[width=0.95\linewidth,trim={0cm 0cm 0cm 0cm},clip]{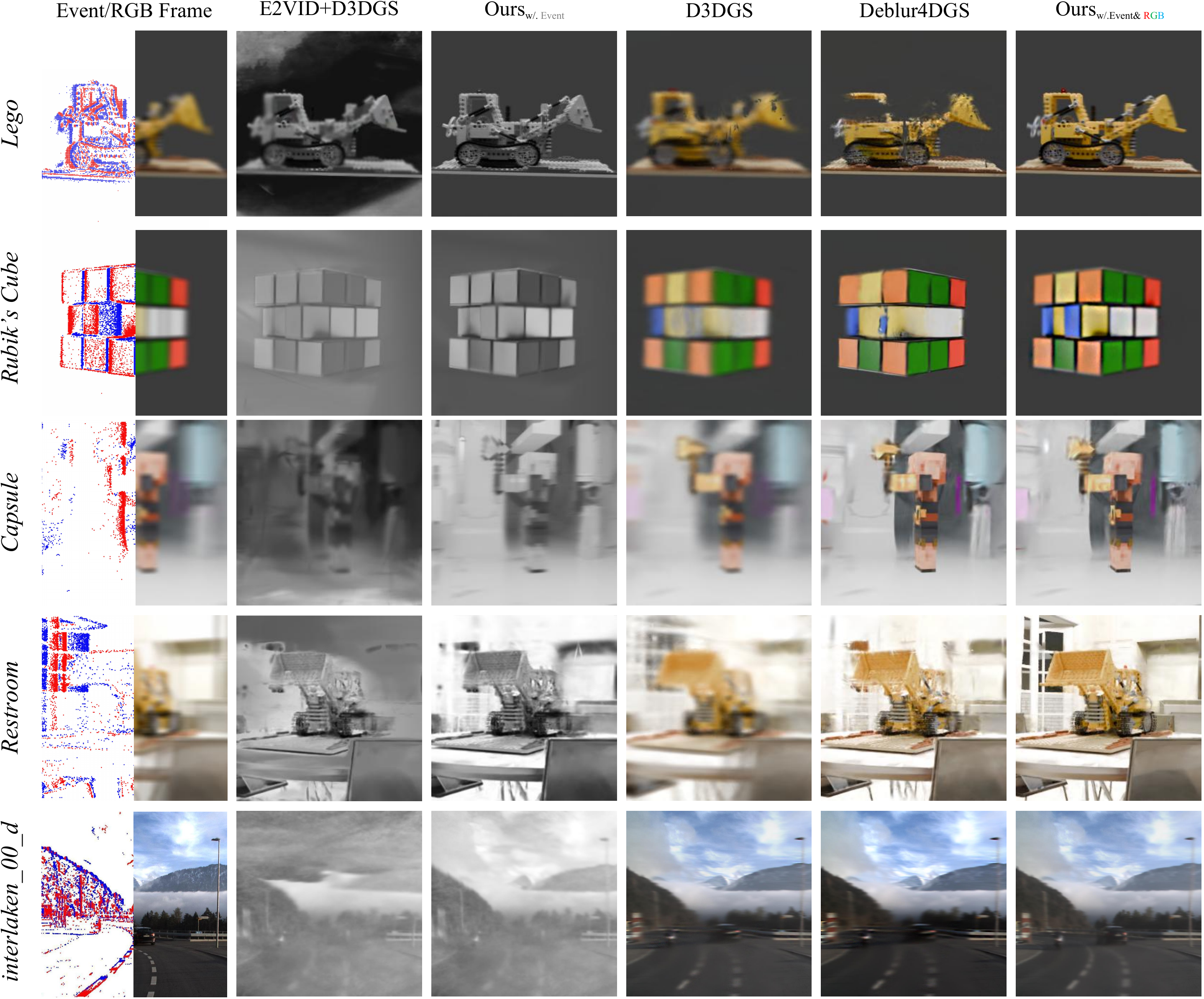}
    \vspace{-1.25em}
    \caption{
         \textbf{Qualitative results of novel view synthesis.} Compared with 4D reconstruction-based methods~\cite{yang2024deformable3dgs,Deblur4DGS}, our approach produces more realistic rendering results with fine-grained details in the synthetic and real scenes. 
    }
    \vspace{-1.0em}
    \label{fig:nvs_evaluation_synthetic}
\end{figure*}

\begin{table*}[!t]
    \centering
    \renewcommand{\arraystretch}{1.25}
    \caption{Quantitative comparison of different methods for novel view synthesis from event streams. The best and second-best results are highlighted in \textbf{bold} and \underline{underlined}, respectively. The average value is computed across 8 synthetic scenes.}
    \vspace{-1.0em}

    \resizebox{\textwidth}{!}{
    \begin{tabular}{l|ccc|ccc|ccc|ccc|ccc}
        \toprule
        \multicolumn{1}{l|}{} & \multicolumn{3}{c|}{Lego} & \multicolumn{3}{c|}{Rubik's Cube} & \multicolumn{3}{c|}{Capsule} & \multicolumn{3}{c|}{Restroom} & \multicolumn{3}{c}{Average} \\
        \multicolumn{1}{l|}{\multirow{-2}{*}{Method}}& ↑PSNR & ↑SSIM & ↓LPIPS & ↑PSNR & ↑SSIM & ↓LPIPS & ↑PSNR & ↑SSIM & ↓LPIPS & ↑PSNR & ↑SSIM & ↓LPIPS & ↑PSNR & ↑SSIM & ↓LPIPS \\
        \midrule
        D3DGS$_{w/o~blur}$ & 26.47 & 0.910 & 0.098 & 20.30 & 0.868 & 0.207 & 31.23 & 0.956 & 0.077 & 28.05 & 0.935 & 0.074 & 23.81 & 0.861 & 0.173 \\
        D3DGS$_{w/~blur}$ & 23.62 & 0.821 & 0.250 & 18.12 & 0.804 & 0.351 & 27.51 & 0.905 & 0.181 & 26.46 & 0.908 & 0.160 & 21.73 & 0.797 & 0.296 \\
        E2VID + D3DGS & 20.57 & 0.765 & 0.347 & 16.16 & 0.752 & 0.404 & 26.06 & 0.851 & 0.268 & 24.87 & 0.856 & 0.247 & 19.88 & 0.728 & 0.397 \\
        Deblur4DGS & 23.17 & 0.813 & 0.265 & 17.68 & 0.786 & 0.375 & 28.06 & 0.908 & 0.176 & 26.35 & 0.900 & 0.162 & 21.66 & 0.797 & 0.291 \\
        \midrule
        E-4DGS$_\textit{\textcolor{gray}{event}-only}$ & \underline{26.85} & \underline{0.912} & \underline{0.084} & \underline{20.97} & \underline{0.882} & \underline{0.185} & \underline{31.85} & \underline{0.959} & \underline{0.071} & \underline{28.83} & \underline{0.942} & \underline{0.069} & \underline{25.38} & \underline{0.896} & \underline{0.134} \\
        \rowcolor{gray!10}
        E-4DGS$_\textit{\textcolor{gray}{event}\& \textcolor{red}{R}\textcolor{green}{G}\textcolor{blue}{B}}$ & \textbf{27.23} & \textbf{0.925} & \textbf{0.078} & \textbf{21.23} & \textbf{0.895} & \textbf{0.172} & \textbf{32.41} & \textbf{0.963} & \textbf{0.068} & \textbf{29.02} & \textbf{0.949} & \textbf{0.067} & \textbf{25.62} & \textbf{0.903} & \textbf{0.129} \\
        \bottomrule
    \end{tabular}
    \vspace{-1.25em}
    }
    \label{tab:nvs_results}
\end{table*}

\begin{table*}[!t]
\centering
\begin{minipage}[t]{0.36\textwidth} 
\centering
\caption{Ablation study of each component.}
\vspace{-0.5em}
\resizebox{\linewidth}{!}{%
\begin{tabular}{ccccc|ccc}
\toprule
\multicolumn{5}{c|}{Method Components} & \multicolumn{3}{c}{Synthetic Datasets} \\
\midrule
$\mathcal{L}_{recon} $ & $\mathcal{L}_{tv} $ & ESS & AES & IIP &  PSNR$\uparrow$ &  SSIM$\uparrow$ &  LPIPS$\downarrow$ \\
\midrule
\rowcolor{gray!10}
\checkmark & \checkmark & \checkmark & \checkmark & \checkmark & \textbf{25.38}  & \textbf{0.896} & \textbf{0.134} \\
\midrule
\checkmark & -- & -- & -- & -- & 23.68  & 0.858 & 0.178 \\
\checkmark & \checkmark & -- & -- & -- & 23.89  & 0.865 & 0.171 \\
\checkmark & \checkmark & \checkmark & -- & -- & 24.71  & 0.876 & 0.153 \\
\checkmark & \checkmark & -- & \checkmark & -- & 23.97  & 0.863 & 0.172 \\
\checkmark & \checkmark & -- & -- & \checkmark & 25.13  & 0.881 & 0.142 \\
\bottomrule
\end{tabular}%
}
\label{tab:ablation_each_component}
\end{minipage}%
\hfill
\begin{minipage}[t]{0.61\textwidth} 
\centering
\caption{Ablation study on the robustness of deblurring. The best and second results are \textbf{bold} and \underline{underlined}, respectively.}
\vspace{-1.3em}

\resizebox{\linewidth}{!}{%
\begin{tabular}{lccc}
\toprule
Blur Degree & Mild blur & Medium blur & Strong blur \\
Metrics & PSNR$\uparrow$/SSIM$\uparrow$/LPIPS$\downarrow$ & PSNR$\uparrow$/SSIM$\uparrow$/LPIPS$\downarrow$ & PSNR$\uparrow$/SSIM$\uparrow$/LPIPS$\downarrow$ \\
\midrule
D3DGS & 19.05 / 0.62 / 0.41 & 18.89 / 0.61 / 0.41 & 16.98 / 0.52 / 0.57 \\
E2VID+D3DGS & 17.79 / 0.55 / 0.49 & 17.69 / 0.54 / 0.49 & 17.93 / 0.55 / 0.49 \\
Deblur4DGS & 19.23 / 0.64 / 0.38 & 18.92 / 0.61 / 0.42 & 16.66 / 0.50 / 0.59 \\
\midrule
E-4DGS$_\textit{\textcolor{gray}{event}-only}$ & \textbf{24.81} / \textbf{0.87} / \textbf{0.17} & \textbf{24.32} / \textbf{0.86} / \textbf{0.19} & \textbf{21.59} / \textbf{0.76} / \textbf{0.28} \\
\rowcolor{gray!10}
E-4DGS$_\textit{\textcolor{gray}{event}\& \textcolor{red}{R}\textcolor{green}{G}\textcolor{blue}{B}}$ & \textbf{24.95} / \textbf{0.88} / \textbf{0.17} & \textbf{24.78} / \textbf{0.87} / \textbf{0.17} & \textbf{22.06} / \textbf{0.80} / \textbf{0.26} \\
\bottomrule
\end{tabular}%
}
\label{tab:abltion:robustness_deblur}
\end{minipage}%
\vspace{-1.5em}

\end{table*}

\subsection{Experimental Evaluation}
\label{sec:5:exp_eval}
\subsubsection{Synthetic dataset}
\label{sec:5:exp_eval:synthetic_dataset}
To generate synthetic data, we render 8 dynamic scenes in Blender~\cite{Blender} at 3000 FPS from six moving viewpoints uniformly distributed around the object at the same height. 
The rendered sequences are then processed by the event simulator v2e~\cite{vid2e} to produce corresponding event streams.

\textbf{(a) Novel View Synthesis:}
As demonstrated in Table~\ref{tab:nvs_results}, our proposed \textit{E-4DGS} outperforms the baselines E2VID + D3DGS across all synthetic scenes in all metrics.
This result is intuitive, as E2VID benefits from being trained on a large dataset but does not account for 3D consistency, whereas our method explicitly incorporates it.
Moreover, EvDNeRF only models the edge of a single object and does not capture the appearance of the dynamic scene, leading to inferior performance compared to the two-stage method and our proposed \textit{E-4DGS}. 
The qualitative comparison of novel view synthesis in Figure~\ref{fig:nvs_evaluation_synthetic} shows that our method produces reconstructed scenes with fewer floaters and more photorealistic rendering results.

\textbf{(b) Motion Blur Decoupling: }
Using event sequences for deblurring blurry RGB frames is a common task. 
In our experiments, we simulate blurry images using Blender~\cite{Blender} by integrating images over the exposure time using LERP and SLERP, which yields realistic, motion-dependent blur.
Table~\ref{tab:nvs_results} show that our method perform better than all 4D reconstruction baselines.
The results of our proposed method are better than the two-stage method which is combining E2VID with frame-based D3DGS.
Furthermore, our method outperforms the frame-based 4D deblurring baseline~\cite{Deblur4DGS}\footnote{This work need to motion masks and frames as inputs. Thus, we utilize MonST3R~\cite{zhang2024monst3r} to extract motion masks and train the whole scenes as the original setting.}, demonstrating that inherent blur-resistant characteristics of events offer greater advantages than relying solely on blur formation.


\textbf{(c) Dynamic Reconstruction with Event and Frame Fusion:}
We combine event sequences and blurry frames by an event-RGB weighted combination, caculated as follows:
\begin{equation}
    \label{eq:loss_event_and_color}
    \mathcal{L}_{Fusion} = \mathcal{L}_{Event} + \lambda_{RGB}\mathcal{L}_{RGB}
\end{equation}


Here, $\mathcal{L}_{RGB}$ is the original photo-realistic rendering loss of D3DGS, including the L1 and D-SSIM loss terms~\cite{yang2024deformable3dgs}. 
Due to the discrete nature of events, while event sequences capture sharp edges, they are noisy in low-light or uniform areas, causing fog-like artifacts in dynamic scenes. 
The color in shaded areas may also be slightly off and requires correction~\cite{low2023robust,rudnev2023eventnerf,rudnev2024dynamic-eventnerf}, as it is inferred from derivative-like data rather than directly measured. 
Incorporating RGB frames addresses these issues by preserving low-frequency texture details while retaining the sharp high-frequency features from event sequences. 
As shown in Figure~\ref{fig:nvs_evaluation_synthetic}, our method reconstructs a sharp dynamic scene with accurate colors, achieving the best performance as reported in Table~\ref{tab:nvs_results}. 
Therefore, color event data from cameras like \textit{DVS346C} is unnecessary, as the predicted color values of event rays can be directly mapped to grayscale.

\begin{figure}[h]
    \centering
    \vspace{-0.75em}
    \includegraphics[width=0.9\linewidth,trim={0cm 0cm 0cm 0cm},clip]{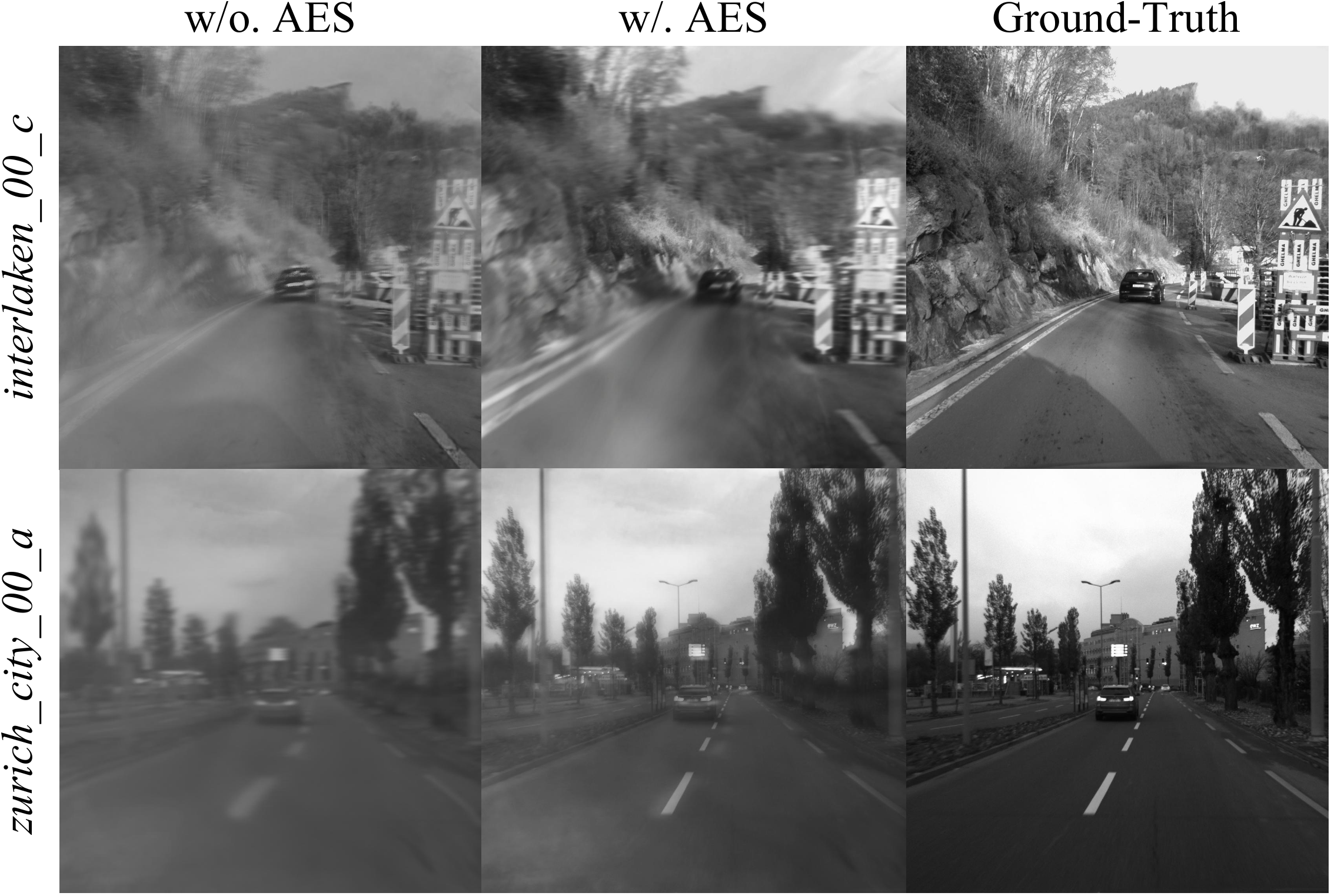}
    \vspace{-0.75em}
    \caption{
        The Performance of the adaptive event supervision on the real-scene of the DSEC dataset.
    }
    \vspace{-1.5em}
    \label{fig:qualitative_AES}
\end{figure}

\subsubsection{Real-scene dataset}
\label{sec:5:exp_eval:real_scene_dataset}
The real-world experiments are conducted on the \textit{interlaken\_00\_c}, \textit{interlaken\_00\_d}, and \textit{zurich\_city\_00\_a} sequences from the autonomous driving dataset DSEC~\cite{Gehrig21ral-DSEC} captured from a modern, high-resolution event sensor—Prophesee Gen3.1.
However, the real-world experiments primarily serve as a qualitative benchmark, as the existing datasets~\cite{peng2024-CoSEC,klenk2021tum-vie,hidalgo2022-EDS}, are not specifically designed for the task of NVS and lacks multi-view event streams with settings comparable to our synthetic dataset. 
This limitation is partly due to the fact that the target novel-view images are captured using a single standard RGB camera, which suffers from saturation effects because of its relatively limited dynamic range. 
Moreover, these images are not raw sensor outputs but have undergone in-camera image processing, often lossy in nature. 
In addition, the spectral response curve of the event camera is not publicly available, making color correction potentially inaccurate when aligning synthesized views with real images.
Consequently, the dataset does not support accurate quantitative NVS evaluation.


For real-scene evaluation of dynamic reconstruction with event and frame fusion, the E2VID + D3DGS baseline recovers more visual details overal. 
However, the proposed E-4DGS produces fewer artifacts, particularly around foreground objects. 
While Deblur4DGS improves on D3DGS, both struggle to recover fine details, such as distant lettering (Figure~\ref{fig:nvs_evaluation_synthetic}). 
E-4DGS outperforms E2VID + D3DGS in high-frequency details and geometry. 
None achieve fully photorealistic quality due to single-view supervision, though such quality is often unnecessary for many robotics applications, emphasizing the need for more multiview event data.

\vspace{-0.8em}
\subsection{Ablation Evaluation}

To assess the contribution of each component, we train various model variants on both synthetic and real sequences, focusing on evaluating effects of event-adaptive slicing splatting (ESS), adaptive event supervision (AES), and intensity importance pruning (IIP).

\textbf{Effect of Different Components.}
For evaluation without ESS, we use a fixed event sampling number for accumulation instead of the specific strategy, and a fixed event threshold value for evaluation without adaptive event supervision. 
As shown in Table~\ref{tab:ablation_each_component}, incorporating ESS and IIP leads to a clear performance improvement. ESS addresses the non-uniform spatial-temporal distribution of event data, while IIP enhances multi-view consistency, together boosting reconstruction performance. 
Although the adaptive event supervision component slightly reduces performance on the synthetic dataset, it significantly improves texture fidelity and temporal consistency in real-world scenarios, as demonstrated in Figure~\ref{fig:qualitative_AES}.

\textbf{Effect of Motion Blur at Different Levels}
To assess the robustness of deblurring, we simulate blurry images with varying degrees of motion blur—mild, medium, and strong—by integrating RGB frames over the exposure time in Blender~\cite{Blender}, creating realistic, motion-dependent blur patterns. 
With the synthetic scene \textit{Garage}, we find that \textit{E-4DGS} consistently outperforms all baselines across all blur levels, achieving the highest performance in Table~\ref{tab:abltion:robustness_deblur}. 
Ours also reconstructs sharper scene details and more accurate object boundaries, particularly under strong blur, demonstrating its superior ability to preserve spatial structure and temporal consistency.





\section{Conclusion}
In this paper, we propose \textbf{E-4DGS}, a novel paradigm for real-time dynamic view synthesis based on dynamic 3DGS using multi-view event sequences.
We design a synthetic multi-view camera setup with six moving event cameras surrounding an object in a 360-degree configuration and provide a benchmark multi-view event stream dataset that captures challenging motion scenarios. 
Our approach outperforms both event-only and event-RGB fusion baselines, paving the way for the exploration of multi-view event-based reconstruction as a novel approach for rapid scene capture.
Future work will focus on addressing the challenges of handling larger-scale dynamic scenes and improving computational efficiency for real-world applications such as autonomous driving and immersive virtual environments.

\clearpage
\section*{Acknowledgement}
This work was supported in part by Natural Science Foundation of China (No.62332002 and No.62202014), and Shenzhen KQTD (No.20240729102051063).




\bibliographystyle{ACM-Reference-Format}
\bibliography{event4dgs}


\begin{thebibliography}{95}


\ifx \showCODEN    \undefined \def \showCODEN     #1{\unskip}     \fi
\ifx \showISBNx    \undefined \def \showISBNx     #1{\unskip}     \fi
\ifx \showISBNxiii \undefined \def \showISBNxiii  #1{\unskip}     \fi
\ifx \showISSN     \undefined \def \showISSN      #1{\unskip}     \fi
\ifx \showLCCN     \undefined \def \showLCCN      #1{\unskip}     \fi
\ifx \shownote     \undefined \def \shownote      #1{#1}          \fi
\ifx \showarticletitle \undefined \def \showarticletitle #1{#1}   \fi
\ifx \showURL      \undefined \def \showURL       {\relax}        \fi
\providecommand\bibfield[2]{#2}
\providecommand\bibinfo[2]{#2}
\providecommand\natexlab[1]{#1}
\providecommand\showeprint[2][]{arXiv:#2}

\bibitem[Bhattacharya et~al\mbox{.}(2024)]%
        {bhattacharya2024evdnerf}
\bibfield{author}{\bibinfo{person}{Anish Bhattacharya}, \bibinfo{person}{Ratnesh Madaan}, \bibinfo{person}{Fernando Cladera}, \bibinfo{person}{Sai Vemprala}, \bibinfo{person}{Rogerio Bonatti}, \bibinfo{person}{Kostas Daniilidis}, \bibinfo{person}{Ashish Kapoor}, \bibinfo{person}{Vijay Kumar}, \bibinfo{person}{Nikolai Matni}, {and} \bibinfo{person}{Jayesh~K Gupta}.} \bibinfo{year}{2024}\natexlab{}.
\newblock \showarticletitle{Evdnerf: Reconstructing event data with dynamic neural radiance fields}. In \bibinfo{booktitle}{\emph{Proceedings of the IEEE/CVF Winter Conference on Applications of Computer Vision}}. \bibinfo{pages}{5846--5855}.
\newblock


\bibitem[Cannici and Scaramuzza(2024)]%
        {evdeblurnerf}
\bibfield{author}{\bibinfo{person}{Marco Cannici} {and} \bibinfo{person}{Davide Scaramuzza}.} \bibinfo{year}{2024}\natexlab{}.
\newblock \showarticletitle{Mitigating Motion Blur in Neural Radiance Fields with Events and Frames}. In \bibinfo{booktitle}{\emph{Proceedings of the IEEE/CVF Conference on Computer Vision and Pattern Recognition (CVPR)}}.
\newblock


\bibitem[Cao and Johnson(2023)]%
        {cao2023hexplane}
\bibfield{author}{\bibinfo{person}{Ang Cao} {and} \bibinfo{person}{Justin Johnson}.} \bibinfo{year}{2023}\natexlab{}.
\newblock \showarticletitle{{HexPlane: A Fast Representation for Dynamic Scenes}}. In \bibinfo{booktitle}{\emph{Computer Vision and Pattern Recognition (CVPR)}}.
\newblock
\urldef\tempurl%
\url{https://caoang327.github.io/HexPlane/}
\showURL{%
\tempurl}


\bibitem[Chen et~al\mbox{.}(2024)]%
        {chen2024usp-gs}
\bibfield{author}{\bibinfo{person}{Kang Chen}, \bibinfo{person}{Jiyuan Zhang}, \bibinfo{person}{Zecheng Hao}, \bibinfo{person}{Yajing Zheng}, \bibinfo{person}{Tiejun Huang}, {and} \bibinfo{person}{Zhaofei Yu}.} \bibinfo{year}{2024}\natexlab{}.
\newblock \showarticletitle{USP-Gaussian: Unifying Spike-based Image Reconstruction, Pose Correction and Gaussian Splatting}.
\newblock \bibinfo{journal}{\emph{arXiv preprint arXiv:2411.10504}} (\bibinfo{year}{2024}).
\newblock


\bibitem[Chen et~al\mbox{.}(2023)]%
        {chen2023clip2scene-understanding}
\bibfield{author}{\bibinfo{person}{Runnan Chen}, \bibinfo{person}{Youquan Liu}, \bibinfo{person}{Lingdong Kong}, \bibinfo{person}{Xinge Zhu}, \bibinfo{person}{Yuexin Ma}, \bibinfo{person}{Yikang Li}, \bibinfo{person}{Yuenan Hou}, \bibinfo{person}{Yu Qiao}, {and} \bibinfo{person}{Wenping Wang}.} \bibinfo{year}{2023}\natexlab{}.
\newblock \showarticletitle{Clip2scene: Towards label-efficient 3d scene understanding by clip}. In \bibinfo{booktitle}{\emph{Proceedings of the IEEE/CVF Conference on Computer Vision and Pattern Recognition}}. \bibinfo{pages}{7020--7030}.
\newblock


\bibitem[Community(2018)]%
        {Blender}
\bibfield{author}{\bibinfo{person}{Blender~Online Community}.} \bibinfo{year}{2018}\natexlab{}.
\newblock \bibinfo{booktitle}{\emph{{Blender - a 3D modelling and rendering package}}}.
\newblock Blender Foundation, Stichting Blender Foundation, Amsterdam.
\newblock
\urldef\tempurl%
\url{http://www.blender.org}
\showURL{%
\tempurl}


\bibitem[Deitke et~al\mbox{.}(2024)]%
        {deitke2024objaverse}
\bibfield{author}{\bibinfo{person}{Matt Deitke}, \bibinfo{person}{Ruoshi Liu}, \bibinfo{person}{Matthew Wallingford}, \bibinfo{person}{Huong Ngo}, \bibinfo{person}{Oscar Michel}, \bibinfo{person}{Aditya Kusupati}, \bibinfo{person}{Alan Fan}, \bibinfo{person}{Christian Laforte}, \bibinfo{person}{Vikram Voleti}, \bibinfo{person}{Samir~Yitzhak Gadre}, {et~al\mbox{.}}} \bibinfo{year}{2024}\natexlab{}.
\newblock \showarticletitle{Objaverse-xl: A universe of 10m+ 3d objects}.
\newblock \bibinfo{journal}{\emph{Advances in Neural Information Processing Systems}}  \bibinfo{volume}{36} (\bibinfo{year}{2024}).
\newblock


\bibitem[Duan et~al\mbox{.}(2024)]%
        {duan20244drotorgs}
\bibfield{author}{\bibinfo{person}{Yuanxing Duan}, \bibinfo{person}{Fangyin Wei}, \bibinfo{person}{Qiyu Dai}, \bibinfo{person}{Yuhang He}, \bibinfo{person}{Wenzheng Chen}, {and} \bibinfo{person}{Baoquan Chen}.} \bibinfo{year}{2024}\natexlab{}.
\newblock \showarticletitle{4d-rotor gaussian splatting: towards efficient novel view synthesis for dynamic scenes}. In \bibinfo{booktitle}{\emph{ACM SIGGRAPH 2024 Conference Papers}}. \bibinfo{pages}{1--11}.
\newblock


\bibitem[Ercan et~al\mbox{.}(2024)]%
        {ercan2024hypere2vid}
\bibfield{author}{\bibinfo{person}{Burak Ercan}, \bibinfo{person}{Onur Eker}, \bibinfo{person}{Canberk Saglam}, \bibinfo{person}{Aykut Erdem}, {and} \bibinfo{person}{Erkut Erdem}.} \bibinfo{year}{2024}\natexlab{}.
\newblock \showarticletitle{Hypere2vid: Improving event-based video reconstruction via hypernetworks}.
\newblock \bibinfo{journal}{\emph{IEEE Transactions on Image Processing}} (\bibinfo{year}{2024}).
\newblock


\bibitem[Fan et~al\mbox{.}(2024)]%
        {fan2024instantsplat}
\bibfield{author}{\bibinfo{person}{Zhiwen Fan}, \bibinfo{person}{Wenyan Cong}, \bibinfo{person}{Kairun Wen}, \bibinfo{person}{Kevin Wang}, \bibinfo{person}{Jian Zhang}, \bibinfo{person}{Xinghao Ding}, \bibinfo{person}{Danfei Xu}, \bibinfo{person}{Boris Ivanovic}, \bibinfo{person}{Marco Pavone}, \bibinfo{person}{Georgios Pavlakos}, \bibinfo{person}{Zhangyang Wang}, {and} \bibinfo{person}{Yue Wang}.} \bibinfo{year}{2024}\natexlab{}.
\newblock \bibinfo{title}{InstantSplat: Unbounded Sparse-view Pose-free Gaussian Splatting in 40 Seconds}.
\newblock
\showeprint[arxiv]{2403.20309}~[cs.CV]


\bibitem[Fan et~al\mbox{.}(2023)]%
        {lightgaussian}
\bibfield{author}{\bibinfo{person}{Zhiwen Fan}, \bibinfo{person}{Kevin Wang}, \bibinfo{person}{Kairun Wen}, \bibinfo{person}{Zehao Zhu}, \bibinfo{person}{Dejia Xu}, {and} \bibinfo{person}{Zhangyang Wang}.} \bibinfo{year}{2023}\natexlab{}.
\newblock \showarticletitle{Lightgaussian: Unbounded 3d gaussian compression with 15x reduction and 200+ fps}.
\newblock \bibinfo{journal}{\emph{arXiv preprint arXiv:2311.17245}} (\bibinfo{year}{2023}).
\newblock


\bibitem[Feng et~al\mbox{.}(2025)]%
        {feng2025ae-nerf}
\bibfield{author}{\bibinfo{person}{Chaoran Feng}, \bibinfo{person}{Wangbo Yu}, \bibinfo{person}{Xinhua Cheng}, \bibinfo{person}{Zhenyu Tang}, \bibinfo{person}{Junwu Zhang}, \bibinfo{person}{Li Yuan}, {and} \bibinfo{person}{Yonghong Tian}.} \bibinfo{year}{2025}\natexlab{}.
\newblock \showarticletitle{AE-NeRF: Augmenting Event-Based Neural Radiance Fields for Non-ideal Conditions and Larger Scene}.
\newblock \bibinfo{journal}{\emph{arXiv preprint arXiv:2501.02807}} (\bibinfo{year}{2025}).
\newblock


\bibitem[Gallego et~al\mbox{.}(2020)]%
        {gallego2020survey}
\bibfield{author}{\bibinfo{person}{Guillermo Gallego}, \bibinfo{person}{Tobi Delbruck}, \bibinfo{person}{Garrick Orchard}, \bibinfo{person}{Chiara Bartolozzi}, \bibinfo{person}{Brian Taba}, \bibinfo{person}{Andrea Censi}, \bibinfo{person}{Stefan Leutenegger}, \bibinfo{person}{Andrew~J Davison}, \bibinfo{person}{Jorg Conradt}, \bibinfo{person}{Kostas Daniilidis}, {et~al\mbox{.}}} \bibinfo{year}{2020}\natexlab{}.
\newblock \showarticletitle{Event-based vision: A survey}.
\newblock \bibinfo{journal}{\emph{IEEE Trans. Pattern Analysis and Machine Intelligence (PAMI)}} (\bibinfo{year}{2020}).
\newblock


\bibitem[Gallego et~al\mbox{.}(2018)]%
        {gallego2018unifying-contrast-maximization}
\bibfield{author}{\bibinfo{person}{Guillermo Gallego}, \bibinfo{person}{Henri Rebecq}, {and} \bibinfo{person}{Davide Scaramuzza}.} \bibinfo{year}{2018}\natexlab{}.
\newblock \showarticletitle{A unifying contrast maximization framework for event cameras, with applications to motion, depth, and optical flow estimation}. In \bibinfo{booktitle}{\emph{Proceedings of the IEEE conference on computer vision and pattern recognition}}. \bibinfo{pages}{3867--3876}.
\newblock


\bibitem[Gehrig et~al\mbox{.}(2021)]%
        {Gehrig21ral-DSEC}
\bibfield{author}{\bibinfo{person}{Mathias Gehrig}, \bibinfo{person}{Willem Aarents}, \bibinfo{person}{Daniel Gehrig}, {and} \bibinfo{person}{Davide Scaramuzza}.} \bibinfo{year}{2021}\natexlab{}.
\newblock \showarticletitle{{DSEC}: A Stereo Event Camera Dataset for Driving Scenarios}.
\newblock \bibinfo{journal}{\emph{{IEEE} Robotics and Automation Letters}} (\bibinfo{year}{2021}).
\newblock
\href{https://doi.org/10.1109/LRA.2021.3068942}{doi:\nolinkurl{10.1109/LRA.2021.3068942}}


\bibitem[Goyal et~al\mbox{.}(2023)]%
        {goyal2023event-hpe-moveenet-cvpr23}
\bibfield{author}{\bibinfo{person}{Gaurvi Goyal}, \bibinfo{person}{Franco Di~Pietro}, \bibinfo{person}{Nicolo Carissimi}, \bibinfo{person}{Arren Glover}, {and} \bibinfo{person}{Chiara Bartolozzi}.} \bibinfo{year}{2023}\natexlab{}.
\newblock \showarticletitle{Moveenet: online high-frequency human pose estimation with an event camera}. In \bibinfo{booktitle}{\emph{Proceedings of the IEEE/CVF Conference on Computer Vision and Pattern Recognition}}. \bibinfo{pages}{4024--4033}.
\newblock


\bibitem[Guo and Gallego(2024)]%
        {guo2024cmax-slam}
\bibfield{author}{\bibinfo{person}{Shuang Guo} {and} \bibinfo{person}{Guillermo Gallego}.} \bibinfo{year}{2024}\natexlab{}.
\newblock \showarticletitle{CMax-SLAM: Event-based rotational-motion bundle adjustment and SLAM system using contrast maximization}.
\newblock \bibinfo{journal}{\emph{IEEE Transactions on Robotics}} (\bibinfo{year}{2024}).
\newblock


\bibitem[Guo et~al\mbox{.}(2024)]%
        {guo2024spikegs}
\bibfield{author}{\bibinfo{person}{Yijia Guo}, \bibinfo{person}{Liwen Hu}, \bibinfo{person}{Yuanxi Bai}, \bibinfo{person}{Jiawei Yao}, \bibinfo{person}{Lei Ma}, {and} \bibinfo{person}{Tiejun Huang}.} \bibinfo{year}{2024}\natexlab{}.
\newblock \showarticletitle{Spikegs: Reconstruct 3d scene via fast-moving bio-inspired sensors}.
\newblock \bibinfo{journal}{\emph{arXiv preprint arXiv:2407.03771}} (\bibinfo{year}{2024}).
\newblock


\bibitem[Han et~al\mbox{.}(2024a)]%
        {han2024event3dgs-nips}
\bibfield{author}{\bibinfo{person}{Haiqian Han}, \bibinfo{person}{Jianing Li}, \bibinfo{person}{Henglu Wei}, {and} \bibinfo{person}{Xiangyang Ji}.} \bibinfo{year}{2024}\natexlab{a}.
\newblock \showarticletitle{Event-3DGS: Event-based 3D Reconstruction Using 3D Gaussian Splatting}.
\newblock \bibinfo{journal}{\emph{Advances in Neural Information Processing Systems}}  \bibinfo{volume}{37} (\bibinfo{year}{2024}), \bibinfo{pages}{128139--128159}.
\newblock


\bibitem[Han et~al\mbox{.}(2024b)]%
        {han2024physical-BECS-simulator}
\bibfield{author}{\bibinfo{person}{Haiqian Han}, \bibinfo{person}{Jiacheng Lyu}, \bibinfo{person}{Jianing Li}, \bibinfo{person}{Henglu Wei}, \bibinfo{person}{Cheng Li}, \bibinfo{person}{Yajing Wei}, \bibinfo{person}{Shu Chen}, {and} \bibinfo{person}{Xiangyang Ji}.} \bibinfo{year}{2024}\natexlab{b}.
\newblock \showarticletitle{Physical-Based Event Camera Simulator}. In \bibinfo{booktitle}{\emph{European Conference on Computer Vision}}. Springer, \bibinfo{pages}{19--35}.
\newblock


\bibitem[He et~al\mbox{.}(2024)]%
        {he2024s4d}
\bibfield{author}{\bibinfo{person}{Bing He}, \bibinfo{person}{Yunuo Chen}, \bibinfo{person}{Guo Lu}, \bibinfo{person}{Qi Wang}, \bibinfo{person}{Qunshan Gu}, \bibinfo{person}{Rong Xie}, \bibinfo{person}{Li Song}, {and} \bibinfo{person}{Wenjun Zhang}.} \bibinfo{year}{2024}\natexlab{}.
\newblock \bibinfo{title}{S4D: Streaming 4D Real-World Reconstruction with Gaussians and 3D Control Points}.
\newblock
\showeprint[arxiv]{2408.13036}~[cs.CV]


\bibitem[He et~al\mbox{.}(2022)]%
        {he2022event-timereplayer}
\bibfield{author}{\bibinfo{person}{Weihua He}, \bibinfo{person}{Kaichao You}, \bibinfo{person}{Zhendong Qiao}, \bibinfo{person}{Xu Jia}, \bibinfo{person}{Ziyang Zhang}, \bibinfo{person}{Wenhui Wang}, \bibinfo{person}{Huchuan Lu}, \bibinfo{person}{Yaoyuan Wang}, {and} \bibinfo{person}{Jianxing Liao}.} \bibinfo{year}{2022}\natexlab{}.
\newblock \showarticletitle{Timereplayer: Unlocking the potential of event cameras for video interpolation}. In \bibinfo{booktitle}{\emph{Proceedings of the IEEE/CVF Conference on Computer Vision and Pattern Recognition}}. \bibinfo{pages}{17804--17813}.
\newblock


\bibitem[Hidalgo-Carri{\'o} et~al\mbox{.}(2022)]%
        {hidalgo2022-EDS}
\bibfield{author}{\bibinfo{person}{Javier Hidalgo-Carri{\'o}}, \bibinfo{person}{Guillermo Gallego}, {and} \bibinfo{person}{Davide Scaramuzza}.} \bibinfo{year}{2022}\natexlab{}.
\newblock \showarticletitle{Event-aided direct sparse odometry}. In \bibinfo{booktitle}{\emph{Proceedings of the IEEE/CVF Conference on Computer Vision and Pattern Recognition}}. \bibinfo{pages}{5781--5790}.
\newblock


\bibitem[Hu et~al\mbox{.}(2021)]%
        {vid2e}
\bibfield{author}{\bibinfo{person}{Yuhuang Hu}, \bibinfo{person}{Shih-Chii Liu}, {and} \bibinfo{person}{Tobi Delbruck}.} \bibinfo{year}{2021}\natexlab{}.
\newblock \showarticletitle{v2e: From video frames to realistic DVS events}. In \bibinfo{booktitle}{\emph{Proceedings of the IEEE/CVF Conference on Computer Vision and Pattern Recognition}}. \bibinfo{pages}{1312--1321}.
\newblock


\bibitem[Huang et~al\mbox{.}(2024)]%
        {huang2024inceventgs}
\bibfield{author}{\bibinfo{person}{Jian Huang}, \bibinfo{person}{Chengrui Dong}, {and} \bibinfo{person}{Peidong Liu}.} \bibinfo{year}{2024}\natexlab{}.
\newblock \showarticletitle{IncEventGS: Pose-Free Gaussian Splatting from a Single Event Camera}.
\newblock \bibinfo{journal}{\emph{arXiv preprint arXiv:2410.08107}} (\bibinfo{year}{2024}).
\newblock


\bibitem[Hwang et~al\mbox{.}(2023)]%
        {hwang2023ev-nerf}
\bibfield{author}{\bibinfo{person}{Inwoo Hwang}, \bibinfo{person}{Junho Kim}, {and} \bibinfo{person}{Young~Min Kim}.} \bibinfo{year}{2023}\natexlab{}.
\newblock \showarticletitle{Ev-nerf: Event based neural radiance field}. In \bibinfo{booktitle}{\emph{Proceedings of the IEEE/CVF Winter Conference on Applications of Computer Vision}}. \bibinfo{pages}{837--847}.
\newblock


\bibitem[Kerbl et~al\mbox{.}(2023)]%
        {kerbl3Dgaussians}
\bibfield{author}{\bibinfo{person}{Bernhard Kerbl}, \bibinfo{person}{Georgios Kopanas}, \bibinfo{person}{Thomas Leimkuhler}, {and} \bibinfo{person}{George Drettakis}.} \bibinfo{year}{2023}\natexlab{}.
\newblock \showarticletitle{3D Gaussian Splatting for Real-Time Radiance Field Rendering}.
\newblock \bibinfo{journal}{\emph{ACM Transactions on Graphics (TOG)}} (\bibinfo{year}{2023}).
\newblock
\urldef\tempurl%
\url{https://repo-sam.inria.fr/fungraph/3d-gaussian-splatting/}
\showURL{%
\tempurl}


\bibitem[Klenk et~al\mbox{.}(2021)]%
        {klenk2021tum-vie}
\bibfield{author}{\bibinfo{person}{Simon Klenk}, \bibinfo{person}{Jason Chui}, \bibinfo{person}{Nikolaus Demmel}, {and} \bibinfo{person}{Daniel Cremers}.} \bibinfo{year}{2021}\natexlab{}.
\newblock \showarticletitle{TUM-VIE: The TUM stereo visual-inertial event dataset}. In \bibinfo{booktitle}{\emph{2021 IEEE/RSJ International Conference on Intelligent Robots and Systems (IROS)}}. IEEE, \bibinfo{pages}{8601--8608}.
\newblock


\bibitem[Klenk et~al\mbox{.}(2023)]%
        {klenk2023e-nerf}
\bibfield{author}{\bibinfo{person}{Simon Klenk}, \bibinfo{person}{Lukas Koestler}, \bibinfo{person}{Davide Scaramuzza}, {and} \bibinfo{person}{Daniel Cremers}.} \bibinfo{year}{2023}\natexlab{}.
\newblock \showarticletitle{E-nerf: Neural radiance fields from a moving event camera}.
\newblock \bibinfo{journal}{\emph{IEEE Robotics and Automation Letters}} \bibinfo{volume}{8}, \bibinfo{number}{3} (\bibinfo{year}{2023}), \bibinfo{pages}{1587--1594}.
\newblock


\bibitem[Lee and Lee(2025)]%
        {lee2025dietgsdiffusionpriorevent}
\bibfield{author}{\bibinfo{person}{Seungjun Lee} {and} \bibinfo{person}{Gim~Hee Lee}.} \bibinfo{year}{2025}\natexlab{}.
\newblock \bibinfo{title}{DiET-GS: Diffusion Prior and Event Stream-Assisted Motion Deblurring 3D Gaussian Splatting}.
\newblock
\showeprint[arxiv]{2503.24210}~[cs.CV]
\urldef\tempurl%
\url{https://arxiv.org/abs/2503.24210}
\showURL{%
\tempurl}


\bibitem[Li et~al\mbox{.}(2023)]%
        {lihao2023freestyleret}
\bibfield{author}{\bibinfo{person}{Hao Li}, \bibinfo{person}{Curise Jia}, \bibinfo{person}{Peng Jin}, \bibinfo{person}{Zesen Cheng}, \bibinfo{person}{Kehan Li}, \bibinfo{person}{Jialu Sui}, \bibinfo{person}{Chang Liu}, {and} \bibinfo{person}{Li Yuan}.} \bibinfo{year}{2023}\natexlab{}.
\newblock \showarticletitle{Freestyleret: Retrieving images from style-diversified queries}.
\newblock \bibinfo{journal}{\emph{arXiv preprint arXiv:2312.02428}} (\bibinfo{year}{2023}).
\newblock


\bibitem[Li et~al\mbox{.}(2025)]%
        {lihao2025decoupled}
\bibfield{author}{\bibinfo{person}{Hao Li}, \bibinfo{person}{Da Long}, \bibinfo{person}{Li Yuan}, \bibinfo{person}{Yu Wang}, \bibinfo{person}{Yonghong Tian}, \bibinfo{person}{Xinchang Wang}, {and} \bibinfo{person}{Fanyang Mo}.} \bibinfo{year}{2025}\natexlab{}.
\newblock \showarticletitle{Decoupled peak property learning for efficient and interpretable electronic circular dichroism spectrum prediction}.
\newblock \bibinfo{journal}{\emph{Nature Computational Science}} (\bibinfo{year}{2025}), \bibinfo{pages}{1--11}.
\newblock


\bibitem[Li* et~al\mbox{.}(2025)]%
        {li2025gs2e}
\bibfield{author}{\bibinfo{person}{Yuchen Li*}, \bibinfo{person}{Chaoran Feng*}, \bibinfo{person}{Zhenyu Tang}, \bibinfo{person}{Kaiyuan Deng}, \bibinfo{person}{Wangbo Yu}, \bibinfo{person}{Yonghong Tian}, {and} \bibinfo{person}{Li Yuan}.} \bibinfo{year}{2025}\natexlab{}.
\newblock \showarticletitle{GS2E: Gaussian Splatting is an Effective Data Generator for Event Stream Generation}.
\newblock \bibinfo{journal}{\emph{arXiv preprint arXiv:2505.15287}} (\bibinfo{year}{2025}).
\newblock


\bibitem[Lin(2024)]%
        {lin2024dynamicnerf-review}
\bibfield{author}{\bibinfo{person}{Jinwei Lin}.} \bibinfo{year}{2024}\natexlab{}.
\newblock \showarticletitle{Dynamic NeRF: A Review}.
\newblock \bibinfo{journal}{\emph{arXiv preprint arXiv:2405.08609}} (\bibinfo{year}{2024}).
\newblock


\bibitem[Lin et~al\mbox{.}(2022)]%
        {lin2022dvs-voltmeter}
\bibfield{author}{\bibinfo{person}{Songnan Lin}, \bibinfo{person}{Ye Ma}, \bibinfo{person}{Zhenhua Guo}, {and} \bibinfo{person}{Bihan Wen}.} \bibinfo{year}{2022}\natexlab{}.
\newblock \showarticletitle{Dvs-voltmeter: Stochastic process-based event simulator for dynamic vision sensors}. In \bibinfo{booktitle}{\emph{European Conference on Computer Vision}}. Springer, \bibinfo{pages}{578--593}.
\newblock


\bibitem[Lin et~al\mbox{.}(2024)]%
        {gaussianflow}
\bibfield{author}{\bibinfo{person}{Youtian Lin}, \bibinfo{person}{Zuozhuo Dai}, \bibinfo{person}{Siyu Zhu}, {and} \bibinfo{person}{Yao Yao}.} \bibinfo{year}{2024}\natexlab{}.
\newblock \showarticletitle{Gaussian-flow: 4d reconstruction with dynamic 3d gaussian particle}. In \bibinfo{booktitle}{\emph{Proceedings of the IEEE/CVF Conference on Computer Vision and Pattern Recognition}}. \bibinfo{pages}{21136--21145}.
\newblock


\bibitem[Low and Lee(2023)]%
        {low2023robust}
\bibfield{author}{\bibinfo{person}{Weng~Fei Low} {and} \bibinfo{person}{Gim~Hee Lee}.} \bibinfo{year}{2023}\natexlab{}.
\newblock \showarticletitle{Robust e-nerf: Nerf from sparse \& noisy events under non-uniform motion}. In \bibinfo{booktitle}{\emph{Proceedings of the IEEE/CVF International Conference on Computer Vision}}.
\newblock


\bibitem[Lu et~al\mbox{.}(2025)]%
        {lu2025dn-4dgs}
\bibfield{author}{\bibinfo{person}{Jiahao Lu}, \bibinfo{person}{Jiacheng Deng}, \bibinfo{person}{Ruijie Zhu}, \bibinfo{person}{Yanzhe Liang}, \bibinfo{person}{Wenfei Yang}, \bibinfo{person}{Xu Zhou}, {and} \bibinfo{person}{Tianzhu Zhang}.} \bibinfo{year}{2025}\natexlab{}.
\newblock \showarticletitle{Dn-4dgs: Denoised deformable network with temporal-spatial aggregation for dynamic scene rendering}.
\newblock \bibinfo{journal}{\emph{Advances in Neural Information Processing Systems}}  \bibinfo{volume}{37} (\bibinfo{year}{2025}), \bibinfo{pages}{84114--84138}.
\newblock


\bibitem[Ma et~al\mbox{.}(2023)]%
        {ma2023de-nerf}
\bibfield{author}{\bibinfo{person}{Qi Ma}, \bibinfo{person}{Danda~Pani Paudel}, \bibinfo{person}{Ajad Chhatkuli}, {and} \bibinfo{person}{Luc Van~Gool}.} \bibinfo{year}{2023}\natexlab{}.
\newblock \showarticletitle{Deformable neural radiance fields using rgb and event cameras}. In \bibinfo{booktitle}{\emph{Proceedings of the IEEE/CVF International Conference on Computer Vision}}. \bibinfo{pages}{3590--3600}.
\newblock


\bibitem[Micusik and Pajdla({[n.\,d.]})]%
        {micusik2006sfm}
\bibfield{author}{\bibinfo{person}{Branislav Micusik} {and} \bibinfo{person}{Tomavs Pajdla}.} \bibinfo{year}{[n.\,d.]}\natexlab{}.
\newblock \showarticletitle{Structure from motion with wide circular field of view cameras}.
\newblock  (\bibinfo{year}{[n.\,d.]}).
\newblock


\bibitem[Mildenhall et~al\mbox{.}(2019)]%
        {mildenhall2019locallightfieldfusion-llff}
\bibfield{author}{\bibinfo{person}{Ben Mildenhall}, \bibinfo{person}{Pratul~P. Srinivasan}, \bibinfo{person}{Rodrigo Ortiz-Cayon}, \bibinfo{person}{Nima~Khademi Kalantari}, \bibinfo{person}{Ravi Ramamoorthi}, \bibinfo{person}{Ren Ng}, {and} \bibinfo{person}{Abhishek Kar}.} \bibinfo{year}{2019}\natexlab{}.
\newblock \showarticletitle{Local Light Field Fusion: Practical View Synthesis with Prescriptive Sampling Guidelines}.
\newblock
\showeprint[arxiv]{1905.00889}~[cs.CV]
\urldef\tempurl%
\url{https://arxiv.org/abs/1905.00889}
\showURL{%
\tempurl}


\bibitem[Mildenhall et~al\mbox{.}(2020)]%
        {nerf}
\bibfield{author}{\bibinfo{person}{Ben Mildenhall}, \bibinfo{person}{Pratul~P Srinivasan}, \bibinfo{person}{Matthew Tancik}, \bibinfo{person}{Jonathan~T Barron}, \bibinfo{person}{Ravi Ramamoorthi}, {and} \bibinfo{person}{Ren Ng}.} \bibinfo{year}{2020}\natexlab{}.
\newblock \showarticletitle{{NeRF: Representing Scenes as Neural Radiance Fields for View Synthesis}}. In \bibinfo{booktitle}{\emph{European Conference on Computer Vision (ECCV)}}.
\newblock
\urldef\tempurl%
\url{https://www.matthewtancik.com/nerf}
\showURL{%
\tempurl}


\bibitem[Pan et~al\mbox{.}(2019)]%
        {pan2019edi}
\bibfield{author}{\bibinfo{person}{Liyuan Pan}, \bibinfo{person}{Cedric Scheerlinck}, \bibinfo{person}{Xin Yu}, \bibinfo{person}{Richard Hartley}, \bibinfo{person}{Miaomiao Liu}, {and} \bibinfo{person}{Yuchao Dai}.} \bibinfo{year}{2019}\natexlab{}.
\newblock \showarticletitle{Bringing a blurry frame alive at high frame-rate with an event camera}. In \bibinfo{booktitle}{\emph{Computer Vision and Pattern Recognition (CVPR)}}.
\newblock


\bibitem[Pang et~al\mbox{.}(2024a)]%
        {pang2024next}
\bibfield{author}{\bibinfo{person}{Yatian Pang}, \bibinfo{person}{Peng Jin}, \bibinfo{person}{Shuo Yang}, \bibinfo{person}{Bin Lin}, \bibinfo{person}{Bin Zhu}, \bibinfo{person}{Zhenyu Tang}, \bibinfo{person}{Liuhan Chen}, \bibinfo{person}{Francis~EH Tay}, \bibinfo{person}{Ser-Nam Lim}, \bibinfo{person}{Harry Yang}, {et~al\mbox{.}}} \bibinfo{year}{2024}\natexlab{a}.
\newblock \showarticletitle{Next patch prediction for autoregressive visual generation}.
\newblock \bibinfo{journal}{\emph{arXiv preprint arXiv:2412.15321}} (\bibinfo{year}{2024}).
\newblock


\bibitem[Pang et~al\mbox{.}(2024b)]%
        {pang2024dreamdance}
\bibfield{author}{\bibinfo{person}{Yatian Pang}, \bibinfo{person}{Bin Zhu}, \bibinfo{person}{Bin Lin}, \bibinfo{person}{Mingzhe Zheng}, \bibinfo{person}{Francis~EH Tay}, \bibinfo{person}{Ser-Nam Lim}, \bibinfo{person}{Harry Yang}, {and} \bibinfo{person}{Li Yuan}.} \bibinfo{year}{2024}\natexlab{b}.
\newblock \showarticletitle{DreamDance: Animating Human Images by Enriching 3D Geometry Cues from 2D Poses}.
\newblock \bibinfo{journal}{\emph{arXiv preprint arXiv:2412.00397}} (\bibinfo{year}{2024}).
\newblock


\bibitem[Park et~al\mbox{.}(2021a)]%
        {park2021nerfies}
\bibfield{author}{\bibinfo{person}{Keunhong Park}, \bibinfo{person}{Utkarsh Sinha}, \bibinfo{person}{Jonathan~T Barron}, \bibinfo{person}{Sofien Bouaziz}, \bibinfo{person}{Dan~B Goldman}, \bibinfo{person}{Steven~M Seitz}, {and} \bibinfo{person}{Ricardo Martin-Brualla}.} \bibinfo{year}{2021}\natexlab{a}.
\newblock \showarticletitle{Nerfies: Deformable neural radiance fields}. In \bibinfo{booktitle}{\emph{Proceedings of the IEEE/CVF International Conference on Computer Vision}}. \bibinfo{pages}{5865--5874}.
\newblock


\bibitem[Park et~al\mbox{.}(2021b)]%
        {park2021hypernerf}
\bibfield{author}{\bibinfo{person}{Keunhong Park}, \bibinfo{person}{Utkarsh Sinha}, \bibinfo{person}{Peter Hedman}, \bibinfo{person}{Jonathan~T Barron}, \bibinfo{person}{Sofien Bouaziz}, \bibinfo{person}{Dan~B Goldman}, \bibinfo{person}{Ricardo Martin-Brualla}, {and} \bibinfo{person}{Steven~M Seitz}.} \bibinfo{year}{2021}\natexlab{b}.
\newblock \showarticletitle{Hypernerf: A higher-dimensional representation for topologically varying neural radiance fields}.
\newblock \bibinfo{journal}{\emph{arXiv preprint arXiv:2106.13228}} (\bibinfo{year}{2021}).
\newblock


\bibitem[Peng et~al\mbox{.}(2024)]%
        {peng2024-CoSEC}
\bibfield{author}{\bibinfo{person}{Shihan Peng}, \bibinfo{person}{Hanyu Zhou}, \bibinfo{person}{Hao Dong}, \bibinfo{person}{Zhiwei Shi}, \bibinfo{person}{Haoyue Liu}, \bibinfo{person}{Yuxing Duan}, \bibinfo{person}{Yi Chang}, {and} \bibinfo{person}{Luxin Yan}.} \bibinfo{year}{2024}\natexlab{}.
\newblock \showarticletitle{CoSEC: A coaxial stereo event camera dataset for autonomous driving}.
\newblock \bibinfo{journal}{\emph{arXiv preprint arXiv:2408.08500}} (\bibinfo{year}{2024}).
\newblock


\bibitem[Pumarola et~al\mbox{.}(2021)]%
        {pumarola2021d-nerf}
\bibfield{author}{\bibinfo{person}{Albert Pumarola}, \bibinfo{person}{Enric Corona}, \bibinfo{person}{Gerard Pons-Moll}, {and} \bibinfo{person}{Francesc Moreno-Noguer}.} \bibinfo{year}{2021}\natexlab{}.
\newblock \showarticletitle{D-nerf: Neural radiance fields for dynamic scenes}. In \bibinfo{booktitle}{\emph{Proceedings of the IEEE/CVF Conference on Computer Vision and Pattern Recognition}}. \bibinfo{pages}{10318--10327}.
\newblock


\bibitem[Qi et~al\mbox{.}(2023)]%
        {qi2023e2nerf}
\bibfield{author}{\bibinfo{person}{Yunshan Qi}, \bibinfo{person}{Lin Zhu}, \bibinfo{person}{Yu Zhang}, {and} \bibinfo{person}{Jia Li}.} \bibinfo{year}{2023}\natexlab{}.
\newblock \showarticletitle{E2NeRF: Event Enhanced Neural Radiance Fields from Blurry Images}. In \bibinfo{booktitle}{\emph{International Conference on Computer Vision (ICCV)}}.
\newblock


\bibitem[Raafat and contributors(2024)]%
        {BlenderNeRF}
\bibfield{author}{\bibinfo{person}{Maxime Raafat} {and} \bibinfo{person}{contributors}.} \bibinfo{year}{2024}\natexlab{}.
\newblock \bibinfo{title}{BlenderNeRF: Easy NeRF synthetic dataset creation within Blender}.
\newblock \bibinfo{howpublished}{\url{https://github.com/maximeraafat/BlenderNeRF}}.
\newblock
\newblock
\shownote{Accessed: 2025-04-07}.


\bibitem[Rebecq et~al\mbox{.}(2019)]%
        {Rebecq19e2vid}
\bibfield{author}{\bibinfo{person}{Henri Rebecq}, \bibinfo{person}{Ren{\'e} Ranftl}, \bibinfo{person}{Vladlen Koltun}, {and} \bibinfo{person}{Davide Scaramuzza}.} \bibinfo{year}{2019}\natexlab{}.
\newblock \showarticletitle{High Speed and High Dynamic Range Video with an Event Camera}.
\newblock \bibinfo{journal}{\emph{{IEEE} Trans. Pattern Anal. Mach. Intell. (T-PAMI)}} (\bibinfo{year}{2019}).
\newblock


\bibitem[Rudnev et~al\mbox{.}(2023)]%
        {rudnev2023eventnerf}
\bibfield{author}{\bibinfo{person}{Viktor Rudnev}, \bibinfo{person}{Mohamed Elgharib}, \bibinfo{person}{Christian Theobalt}, {and} \bibinfo{person}{Vladislav Golyanik}.} \bibinfo{year}{2023}\natexlab{}.
\newblock \showarticletitle{EventNeRF: Neural Radiance Fields from a Single Colour Event Camera}. In \bibinfo{booktitle}{\emph{Computer Vision and Pattern Recognition (CVPR)}}.
\newblock


\bibitem[Rudnev et~al\mbox{.}(2024)]%
        {rudnev2024dynamic-eventnerf}
\bibfield{author}{\bibinfo{person}{Viktor Rudnev}, \bibinfo{person}{Gereon Fox}, \bibinfo{person}{Mohamed Elgharib}, \bibinfo{person}{Christian Theobalt}, {and} \bibinfo{person}{Vladislav Golyanik}.} \bibinfo{year}{2024}\natexlab{}.
\newblock \showarticletitle{Dynamic EventNeRF: Reconstructing General Dynamic Scenes from Multi-view Event Cameras}.
\newblock \bibinfo{journal}{\emph{arXiv preprint arXiv:2412.06770}} (\bibinfo{year}{2024}).
\newblock


\bibitem[Savran et~al\mbox{.}(2018)]%
        {savran2018-event-lip-energy}
\bibfield{author}{\bibinfo{person}{Arman Savran}, \bibinfo{person}{Raffaele Tavarone}, \bibinfo{person}{Bertrand Higy}, \bibinfo{person}{Leonardo Badino}, {and} \bibinfo{person}{Chiara Bartolozzi}.} \bibinfo{year}{2018}\natexlab{}.
\newblock \showarticletitle{Energy and computation efficient audio-visual voice activity detection driven by event-cameras}. In \bibinfo{booktitle}{\emph{2018 13th IEEE International Conference on Automatic Face \& Gesture Recognition (FG 2018)}}. IEEE, \bibinfo{pages}{333--340}.
\newblock


\bibitem[Schonberger and Frahm(2016)]%
        {colmap}
\bibfield{author}{\bibinfo{person}{Johannes~L Schonberger} {and} \bibinfo{person}{Jan-Michael Frahm}.} \bibinfo{year}{2016}\natexlab{}.
\newblock \showarticletitle{{Structure-from-motion Revisited}}. In \bibinfo{booktitle}{\emph{Computer Vision and Pattern Recognition (CVPR)}}.
\newblock


\bibitem[Shan et~al\mbox{.}(2025)]%
        {shan2025deformablegs-efficient-icra25}
\bibfield{author}{\bibinfo{person}{Jiwei Shan}, \bibinfo{person}{Zeyu Cai}, \bibinfo{person}{Cheng-Tai Hsieh}, \bibinfo{person}{Shing~Shin Cheng}, {and} \bibinfo{person}{Hesheng Wang}.} \bibinfo{year}{2025}\natexlab{}.
\newblock \showarticletitle{Deformable Gaussian Splatting for Efficient and High-Fidelity Reconstruction of Surgical Scenes}.
\newblock \bibinfo{journal}{\emph{arXiv preprint arXiv:2501.01101}} (\bibinfo{year}{2025}).
\newblock


\bibitem[Shao et~al\mbox{.}(2023)]%
        {eicil_nips_23}
\bibfield{author}{\bibinfo{person}{Zihang Shao}, \bibinfo{person}{Xuanye Fang}, \bibinfo{person}{Yaxin Li}, \bibinfo{person}{Chaoran Feng}, \bibinfo{person}{Jiangrong Shen}, {and} \bibinfo{person}{Qi Xu}.} \bibinfo{year}{2023}\natexlab{}.
\newblock \showarticletitle{EICIL: joint excitatory inhibitory cycle iteration learning for deep spiking neural networks}.
\newblock \bibinfo{journal}{\emph{Advances in Neural Information Processing Systems}}  \bibinfo{volume}{36} (\bibinfo{year}{2023}), \bibinfo{pages}{32117--32128}.
\newblock


\bibitem[Snavely et~al\mbox{.}(2006)]%
        {snavely2006photo-tourism}
\bibfield{author}{\bibinfo{person}{Noah Snavely}, \bibinfo{person}{Steven~M Seitz}, {and} \bibinfo{person}{Richard Szeliski}.} \bibinfo{year}{2006}\natexlab{}.
\newblock \showarticletitle{Photo tourism: exploring photo collections in 3D}.
\newblock In \bibinfo{booktitle}{\emph{ACM siggraph 2006 papers}}. \bibinfo{pages}{835--846}.
\newblock


\bibitem[Tan et~al\mbox{.}(2022)]%
        {tan2022event-spatio-temporal-lip-cvpr22}
\bibfield{author}{\bibinfo{person}{Ganchao Tan}, \bibinfo{person}{Yang Wang}, \bibinfo{person}{Han Han}, \bibinfo{person}{Yang Cao}, \bibinfo{person}{Feng Wu}, {and} \bibinfo{person}{Zheng-Jun Zha}.} \bibinfo{year}{2022}\natexlab{}.
\newblock \showarticletitle{Multi-grained spatio-temporal features perceived network for event-based lip-reading}. In \bibinfo{booktitle}{\emph{Proceedings of the IEEE/CVF Conference on Computer Vision and Pattern Recognition}}. \bibinfo{pages}{20094--20103}.
\newblock


\bibitem[Tang et~al\mbox{.}(2024a)]%
        {tang2024lse-nerf}
\bibfield{author}{\bibinfo{person}{Wei~Zhi Tang}, \bibinfo{person}{Daniel Rebain}, \bibinfo{person}{Kostantinos~G Derpanis}, {and} \bibinfo{person}{Kwang~Moo Yi}.} \bibinfo{year}{2024}\natexlab{a}.
\newblock \showarticletitle{LSE-NeRF: Learning Sensor Modeling Errors for Deblured Neural Radiance Fields with RGB-Event Stereo}.
\newblock \bibinfo{journal}{\emph{arXiv preprint arXiv:2409.06104}} (\bibinfo{year}{2024}).
\newblock


\bibitem[Tang et~al\mbox{.}(2025)]%
        {tang2025neuralgs}
\bibfield{author}{\bibinfo{person}{Zhenyu Tang}, \bibinfo{person}{Chaoran Feng}, \bibinfo{person}{Xinhua Cheng}, \bibinfo{person}{Wangbo Yu}, \bibinfo{person}{Junwu Zhang}, \bibinfo{person}{Yuan Liu}, \bibinfo{person}{Xiaoxiao Long}, \bibinfo{person}{Wenping Wang}, {and} \bibinfo{person}{Li Yuan}.} \bibinfo{year}{2025}\natexlab{}.
\newblock \showarticletitle{NeuralGS: Bridging Neural Fields and 3D Gaussian Splatting for Compact 3D Representations}.
\newblock \bibinfo{journal}{\emph{arXiv preprint arXiv:2503.23162}} (\bibinfo{year}{2025}).
\newblock


\bibitem[Tang et~al\mbox{.}(2024b)]%
        {tang2024cycle3d}
\bibfield{author}{\bibinfo{person}{Zhenyu Tang}, \bibinfo{person}{Junwu Zhang}, \bibinfo{person}{Xinhua Cheng}, \bibinfo{person}{Wangbo Yu}, \bibinfo{person}{Chaoran Feng}, \bibinfo{person}{Yatian Pang}, \bibinfo{person}{Bin Lin}, {and} \bibinfo{person}{Li Yuan}.} \bibinfo{year}{2024}\natexlab{b}.
\newblock \showarticletitle{Cycle3D: High-quality and Consistent Image-to-3D Generation via Generation-Reconstruction Cycle}.
\newblock \bibinfo{journal}{\emph{arXiv preprint arXiv:2407.19548}} (\bibinfo{year}{2024}).
\newblock


\bibitem[Tulyakov et~al\mbox{.}(2021)]%
        {tulyakov2021event-time-lens}
\bibfield{author}{\bibinfo{person}{Stepan Tulyakov}, \bibinfo{person}{Daniel Gehrig}, \bibinfo{person}{Stamatios Georgoulis}, \bibinfo{person}{Julius Erbach}, \bibinfo{person}{Mathias Gehrig}, \bibinfo{person}{Yuanyou Li}, {and} \bibinfo{person}{Davide Scaramuzza}.} \bibinfo{year}{2021}\natexlab{}.
\newblock \showarticletitle{Time lens: Event-based video frame interpolation}. In \bibinfo{booktitle}{\emph{Proceedings of the IEEE/CVF conference on computer vision and pattern recognition}}. \bibinfo{pages}{16155--16164}.
\newblock


\bibitem[Wan et~al\mbox{.}(2024)]%
        {wan2024template-free-4dgs-nips2024}
\bibfield{author}{\bibinfo{person}{Diwen Wan}, \bibinfo{person}{Yuxiang Wang}, \bibinfo{person}{Ruijie Lu}, {and} \bibinfo{person}{Gang Zeng}.} \bibinfo{year}{2024}\natexlab{}.
\newblock \showarticletitle{Template-free Articulated Gaussian Splatting for Real-time Reposable Dynamic View Synthesis}.
\newblock \bibinfo{journal}{\emph{arXiv preprint arXiv:2412.05570}} (\bibinfo{year}{2024}).
\newblock


\bibitem[Wang et~al\mbox{.}(2025)]%
        {wang2025-Event-Cameras-High-mobility-Devices-Survey}
\bibfield{author}{\bibinfo{person}{Haoyang Wang}, \bibinfo{person}{Ruishan Guo}, \bibinfo{person}{Pengtao Ma}, \bibinfo{person}{Ciyu Ruan}, \bibinfo{person}{Xinyu Luo}, \bibinfo{person}{Wenhua Ding}, \bibinfo{person}{Tianyang Zhong}, \bibinfo{person}{Jingao Xu}, \bibinfo{person}{Yunhao Liu}, {and} \bibinfo{person}{Xinlei Chen}.} \bibinfo{year}{2025}\natexlab{}.
\newblock \showarticletitle{Towards Mobile Sensing with Event Cameras on High-mobility Resource-constrained Devices: A Survey}.
\newblock \bibinfo{journal}{\emph{arXiv preprint arXiv:2503.22943}} (\bibinfo{year}{2025}).
\newblock


\bibitem[Wang et~al\mbox{.}(2024a)]%
        {wang2024evggs}
\bibfield{author}{\bibinfo{person}{Jiaxu Wang}, \bibinfo{person}{Junhao He}, \bibinfo{person}{Ziyi Zhang}, \bibinfo{person}{Mingyuan Sun}, \bibinfo{person}{Jingkai Sun}, {and} \bibinfo{person}{Renjing Xu}.} \bibinfo{year}{2024}\natexlab{a}.
\newblock \showarticletitle{EvGGS: A Collaborative Learning Framework for Event-based Generalizable Gaussian Splatting}.
\newblock \bibinfo{journal}{\emph{arXiv preprint arXiv:2405.14959}} (\bibinfo{year}{2024}).
\newblock


\bibitem[Wang et~al\mbox{.}(2024b)]%
        {wang2024shape-of-motion}
\bibfield{author}{\bibinfo{person}{Qianqian Wang}, \bibinfo{person}{Vickie Ye}, \bibinfo{person}{Hang Gao}, \bibinfo{person}{Jake Austin}, \bibinfo{person}{Zhengqi Li}, {and} \bibinfo{person}{Angjoo Kanazawa}.} \bibinfo{year}{2024}\natexlab{b}.
\newblock \showarticletitle{Shape of motion: 4d reconstruction from a single video}.
\newblock \bibinfo{journal}{\emph{arXiv preprint arXiv:2407.13764}} (\bibinfo{year}{2024}).
\newblock


\bibitem[Wu et~al\mbox{.}(2025)]%
        {wu2025swift4dgs}
\bibfield{author}{\bibinfo{person}{Jiahao Wu}, \bibinfo{person}{Rui Peng}, \bibinfo{person}{Zhiyan Wang}, \bibinfo{person}{Lu Xiao}, \bibinfo{person}{Luyang Tang}, \bibinfo{person}{Jinbo Yan}, \bibinfo{person}{Kaiqiang Xiong}, {and} \bibinfo{person}{Ronggang Wang}.} \bibinfo{year}{2025}\natexlab{}.
\newblock \showarticletitle{Swift4D: Adaptive divide-and-conquer Gaussian Splatting for compact and efficient reconstruction of dynamic scene}.
\newblock \bibinfo{journal}{\emph{arXiv preprint arXiv:2503.12307}} (\bibinfo{year}{2025}).
\newblock


\bibitem[Wu et~al\mbox{.}(2024c)]%
        {wu2024ev-gs}
\bibfield{author}{\bibinfo{person}{Jingqian Wu}, \bibinfo{person}{Shuo Zhu}, \bibinfo{person}{Chutian Wang}, {and} \bibinfo{person}{Edmund~Y Lam}.} \bibinfo{year}{2024}\natexlab{c}.
\newblock \showarticletitle{Ev-GS: Event-based gaussian splatting for efficient and accurate radiance field rendering}. In \bibinfo{booktitle}{\emph{2024 IEEE 34th International Workshop on Machine Learning for Signal Processing (MLSP)}}. IEEE, \bibinfo{pages}{1--6}.
\newblock


\bibitem[Wu et~al\mbox{.}(2024d)]%
        {wu2024-sweepevgs}
\bibfield{author}{\bibinfo{person}{Jingqian Wu}, \bibinfo{person}{Shuo Zhu}, \bibinfo{person}{Chutian Wang}, \bibinfo{person}{Boxin Shi}, {and} \bibinfo{person}{Edmund~Y Lam}.} \bibinfo{year}{2024}\natexlab{d}.
\newblock \showarticletitle{SweepEvGS: Event-Based 3D Gaussian Splatting for Macro and Micro Radiance Field Rendering from a Single Sweep}.
\newblock \bibinfo{journal}{\emph{arXiv preprint arXiv:2412.11579}} (\bibinfo{year}{2024}).
\newblock


\bibitem[Wu et~al\mbox{.}(2024b)]%
        {Deblur4DGS}
\bibfield{author}{\bibinfo{person}{Renlong Wu}, \bibinfo{person}{Zhilu Zhang}, \bibinfo{person}{Mingyang Chen}, \bibinfo{person}{Xiaopeng Fan}, \bibinfo{person}{Zifei Yan}, {and} \bibinfo{person}{Wangmeng Zuo}.} \bibinfo{year}{2024}\natexlab{b}.
\newblock \showarticletitle{Deblur4DGS: 4D Gaussian Splatting from Blurry Monocular Video}.
\newblock \bibinfo{journal}{\emph{arXiv preprint arXiv:2412.06424}} (\bibinfo{year}{2024}).
\newblock


\bibitem[Wu et~al\mbox{.}(2024a)]%
        {wu2024event-leod-detection-cvpr24}
\bibfield{author}{\bibinfo{person}{Ziyi Wu}, \bibinfo{person}{Mathias Gehrig}, \bibinfo{person}{Qing Lyu}, \bibinfo{person}{Xudong Liu}, {and} \bibinfo{person}{Igor Gilitschenski}.} \bibinfo{year}{2024}\natexlab{a}.
\newblock \showarticletitle{Leod: Label-efficient object detection for event cameras}. In \bibinfo{booktitle}{\emph{Proceedings of the IEEE/CVF Conference on Computer Vision and Pattern Recognition}}. \bibinfo{pages}{16933--16943}.
\newblock


\bibitem[Xiong et~al\mbox{.}(2024)]%
        {xiong2024event3dgs}
\bibfield{author}{\bibinfo{person}{Tianyi Xiong}, \bibinfo{person}{Jiayi Wu}, \bibinfo{person}{Botao He}, \bibinfo{person}{Cornelia Fermuller}, \bibinfo{person}{Yiannis Aloimonos}, \bibinfo{person}{Heng Huang}, {and} \bibinfo{person}{Christopher~A Metzler}.} \bibinfo{year}{2024}\natexlab{}.
\newblock \showarticletitle{Event3DGS: Event-based 3D Gaussian Splatting for Fast Egomotion}.
\newblock \bibinfo{journal}{\emph{arXiv preprint arXiv:2406.02972}} (\bibinfo{year}{2024}).
\newblock


\bibitem[Xu et~al\mbox{.}(2025)]%
        {xu2025-event-driven-3d-reconstrution-survey}
\bibfield{author}{\bibinfo{person}{Chuanzhi Xu}, \bibinfo{person}{Haoxian Zhou}, \bibinfo{person}{Haodong Chen}, \bibinfo{person}{Vera Chung}, {and} \bibinfo{person}{Qiang Qu}.} \bibinfo{year}{2025}\natexlab{}.
\newblock \showarticletitle{A Survey on Event-driven 3D Reconstruction: Development under Different Categories}.
\newblock \bibinfo{journal}{\emph{arXiv preprint arXiv:2503.19753}} (\bibinfo{year}{2025}).
\newblock


\bibitem[Xu et~al\mbox{.}(2024)]%
        {xu2024event-boosted}
\bibfield{author}{\bibinfo{person}{Wenhao Xu}, \bibinfo{person}{Wenming Weng}, \bibinfo{person}{Yueyi Zhang}, \bibinfo{person}{Ruikang Xu}, {and} \bibinfo{person}{Zhiwei Xiong}.} \bibinfo{year}{2024}\natexlab{}.
\newblock \showarticletitle{Event-boosted Deformable 3D Gaussians for Fast Dynamic Scene Reconstruction}.
\newblock \bibinfo{journal}{\emph{arXiv preprint arXiv:2411.16180}} (\bibinfo{year}{2024}).
\newblock


\bibitem[Xue et~al\mbox{.}(2023)]%
        {xue2023ulip-3d-pcl-understanding}
\bibfield{author}{\bibinfo{person}{Le Xue}, \bibinfo{person}{Mingfei Gao}, \bibinfo{person}{Chen Xing}, \bibinfo{person}{Roberto Mart{\'\i}n-Mart{\'\i}n}, \bibinfo{person}{Jiajun Wu}, \bibinfo{person}{Caiming Xiong}, \bibinfo{person}{Ran Xu}, \bibinfo{person}{Juan~Carlos Niebles}, {and} \bibinfo{person}{Silvio Savarese}.} \bibinfo{year}{2023}\natexlab{}.
\newblock \showarticletitle{Ulip: Learning a unified representation of language, images, and point clouds for 3d understanding}. In \bibinfo{booktitle}{\emph{Proceedings of the IEEE/CVF conference on computer vision and pattern recognition}}. \bibinfo{pages}{1179--1189}.
\newblock


\bibitem[Yan et~al\mbox{.}(2024b)]%
        {yan2024gs-slam}
\bibfield{author}{\bibinfo{person}{Chi Yan}, \bibinfo{person}{Delin Qu}, \bibinfo{person}{Dan Xu}, \bibinfo{person}{Bin Zhao}, \bibinfo{person}{Zhigang Wang}, \bibinfo{person}{Dong Wang}, {and} \bibinfo{person}{Xuelong Li}.} \bibinfo{year}{2024}\natexlab{b}.
\newblock \showarticletitle{Gs-slam: Dense visual slam with 3d gaussian splatting}. In \bibinfo{booktitle}{\emph{Proceedings of the IEEE/CVF Conference on Computer Vision and Pattern Recognition}}.
\newblock


\bibitem[Yan et~al\mbox{.}(2024a)]%
        {yan2024-4dgs-Scale-aware-Residual-Field}
\bibfield{author}{\bibinfo{person}{Jinbo Yan}, \bibinfo{person}{Rui Peng}, \bibinfo{person}{Luyang Tang}, {and} \bibinfo{person}{Ronggang Wang}.} \bibinfo{year}{2024}\natexlab{a}.
\newblock \showarticletitle{4D Gaussian Splatting with Scale-aware Residual Field and Adaptive Optimization for Real-time rendering of temporally complex dynamic scenes}. In \bibinfo{booktitle}{\emph{Proceedings of the 32nd ACM International Conference on Multimedia}}. \bibinfo{pages}{7871--7880}.
\newblock


\bibitem[Yan et~al\mbox{.}(2023)]%
        {yan2023nerf-ds}
\bibfield{author}{\bibinfo{person}{Zhiwen Yan}, \bibinfo{person}{Chen Li}, {and} \bibinfo{person}{Gim~Hee Lee}.} \bibinfo{year}{2023}\natexlab{}.
\newblock \showarticletitle{Nerf-ds: Neural radiance fields for dynamic specular objects}. In \bibinfo{booktitle}{\emph{Proceedings of the IEEE/CVF Conference on Computer Vision and Pattern Recognition}}. \bibinfo{pages}{8285--8295}.
\newblock


\bibitem[Yang et~al\mbox{.}(2024a)]%
        {yang2024deformable3dgs}
\bibfield{author}{\bibinfo{person}{Ziyi Yang}, \bibinfo{person}{Xinyu Gao}, \bibinfo{person}{Wen Zhou}, \bibinfo{person}{Shaohui Jiao}, \bibinfo{person}{Yuqing Zhang}, {and} \bibinfo{person}{Xiaogang Jin}.} \bibinfo{year}{2024}\natexlab{a}.
\newblock \showarticletitle{Deformable 3d gaussians for high-fidelity monocular dynamic scene reconstruction}. In \bibinfo{booktitle}{\emph{Proceedings of the IEEE/CVF Conference on Computer Vision and Pattern Recognition}}. \bibinfo{pages}{20331--20341}.
\newblock


\bibitem[Yang et~al\mbox{.}(2024b)]%
        {yang20244dgs-iclr2024}
\bibfield{author}{\bibinfo{person}{Zeyu Yang}, \bibinfo{person}{Zijie Pan}, \bibinfo{person}{Xiatian Zhu}, \bibinfo{person}{Li Zhang}, \bibinfo{person}{Yu-Gang Jiang}, {and} \bibinfo{person}{Philip~HS Torr}.} \bibinfo{year}{2024}\natexlab{b}.
\newblock \showarticletitle{4D Gaussian Splatting: Modeling Dynamic Scenes with Native 4D Primitives}.
\newblock \bibinfo{journal}{\emph{arXiv preprint arXiv:2412.20720}} (\bibinfo{year}{2024}).
\newblock


\bibitem[Yin et~al\mbox{.}(2024)]%
        {yin2024e-3dgs}
\bibfield{author}{\bibinfo{person}{Xiaoting Yin}, \bibinfo{person}{Hao Shi}, \bibinfo{person}{Yuhan Bao}, \bibinfo{person}{Zhenshan Bing}, \bibinfo{person}{Yiyi Liao}, \bibinfo{person}{Kailun Yang}, {and} \bibinfo{person}{Kaiwei Wang}.} \bibinfo{year}{2024}\natexlab{}.
\newblock \showarticletitle{E-3DGS: Gaussian Splatting with Exposure and Motion Events}.
\newblock \bibinfo{journal}{\emph{arXiv preprint arXiv:2410.16995}} (\bibinfo{year}{2024}).
\newblock


\bibitem[YU et~al\mbox{.}(2025)]%
        {yu2025trajectorycrafter}
\bibfield{author}{\bibinfo{person}{Mark YU}, \bibinfo{person}{Wenbo Hu}, \bibinfo{person}{Jinbo Xing}, {and} \bibinfo{person}{Ying Shan}.} \bibinfo{year}{2025}\natexlab{}.
\newblock \showarticletitle{TrajectoryCrafter: Redirecting Camera Trajectory for Monocular Videos via Diffusion Models}.
\newblock \bibinfo{journal}{\emph{arXiv preprint arXiv:2503.05638}} (\bibinfo{year}{2025}).
\newblock


\bibitem[Yu et~al\mbox{.}(2024b)]%
        {evagaussians}
\bibfield{author}{\bibinfo{person}{Wangbo Yu}, \bibinfo{person}{Chaoran Feng}, \bibinfo{person}{Jiye Tang}, \bibinfo{person}{Xu Jia}, \bibinfo{person}{Li Yuan}, {and} \bibinfo{person}{Yonghong Tian}.} \bibinfo{year}{2024}\natexlab{b}.
\newblock \showarticletitle{EvaGaussians: Event Stream Assisted Gaussian Splatting from Blurry Images}.
\newblock \bibinfo{journal}{\emph{arXiv preprint arXiv:2405.20224}} (\bibinfo{year}{2024}).
\newblock


\bibitem[Yu et~al\mbox{.}(2024c)]%
        {yu2024viewcrafter}
\bibfield{author}{\bibinfo{person}{Wangbo Yu}, \bibinfo{person}{Jinbo Xing}, \bibinfo{person}{Li Yuan}, \bibinfo{person}{Wenbo Hu}, \bibinfo{person}{Xiaoyu Li}, \bibinfo{person}{Zhipeng Huang}, \bibinfo{person}{Xiangjun Gao}, \bibinfo{person}{Tien-Tsin Wong}, \bibinfo{person}{Ying Shan}, {and} \bibinfo{person}{Yonghong Tian}.} \bibinfo{year}{2024}\natexlab{c}.
\newblock \showarticletitle{Viewcrafter: Taming video diffusion models for high-fidelity novel view synthesis}.
\newblock \bibinfo{journal}{\emph{arXiv preprint arXiv:2409.02048}} (\bibinfo{year}{2024}).
\newblock


\bibitem[Yu et~al\mbox{.}(2024a)]%
        {mip-splatting-cvpr24}
\bibfield{author}{\bibinfo{person}{Zehao Yu}, \bibinfo{person}{Anpei Chen}, \bibinfo{person}{Binbin Huang}, \bibinfo{person}{Torsten Sattler}, {and} \bibinfo{person}{Andreas Geiger}.} \bibinfo{year}{2024}\natexlab{a}.
\newblock \showarticletitle{Mip-splatting: Alias-free 3d gaussian splatting}. In \bibinfo{booktitle}{\emph{Proceedings of the IEEE/CVF Conference on Computer Vision and Pattern Recognition}}. \bibinfo{pages}{19447--19456}.
\newblock


\bibitem[Yuan et~al\mbox{.}(2024)]%
        {yuanshenghai2024magictime}
\bibfield{author}{\bibinfo{person}{Shenghai Yuan}, \bibinfo{person}{Jinfa Huang}, \bibinfo{person}{Yujun Shi}, \bibinfo{person}{Yongqi Xu}, \bibinfo{person}{Ruijie Zhu}, \bibinfo{person}{Bin Lin}, \bibinfo{person}{Xinhua Cheng}, \bibinfo{person}{Li Yuan}, {and} \bibinfo{person}{Jiebo Luo}.} \bibinfo{year}{2024}\natexlab{}.
\newblock \showarticletitle{MagicTime: Time-lapse Video Generation Models as Metamorphic Simulators}.
\newblock \bibinfo{journal}{\emph{arXiv preprint arXiv:2404.05014}} (\bibinfo{year}{2024}).
\newblock


\bibitem[Zahid et~al\mbox{.}(2025)]%
        {zahid2025e3dgs-3dv}
\bibfield{author}{\bibinfo{person}{Sohaib Zahid}, \bibinfo{person}{Viktor Rudnev}, \bibinfo{person}{Eddy Ilg}, {and} \bibinfo{person}{Vladislav Golyanik}.} \bibinfo{year}{2025}\natexlab{}.
\newblock \showarticletitle{E-3DGS: Event-based Novel View Rendering of Large-scale Scenes Using 3D Gaussian Splatting}.
\newblock \bibinfo{journal}{\emph{3DV}} (\bibinfo{year}{2025}).
\newblock


\bibitem[Zhang et~al\mbox{.}(2024c)]%
        {zhang2024monst3r}
\bibfield{author}{\bibinfo{person}{Junyi Zhang}, \bibinfo{person}{Charles Herrmann}, \bibinfo{person}{Junhwa Hur}, \bibinfo{person}{Varun Jampani}, \bibinfo{person}{Trevor Darrell}, \bibinfo{person}{Forrester Cole}, \bibinfo{person}{Deqing Sun}, {and} \bibinfo{person}{Ming-Hsuan Yang}.} \bibinfo{year}{2024}\natexlab{c}.
\newblock \showarticletitle{Monst3r: A simple approach for estimating geometry in the presence of motion}.
\newblock \bibinfo{journal}{\emph{arXiv preprint arXiv:2410.03825}} (\bibinfo{year}{2024}).
\newblock


\bibitem[Zhang et~al\mbox{.}(2023)]%
        {zhang2023repaint123}
\bibfield{author}{\bibinfo{person}{Junwu Zhang}, \bibinfo{person}{Zhenyu Tang}, \bibinfo{person}{Yatian Pang}, \bibinfo{person}{Xinhua Cheng}, \bibinfo{person}{Peng Jin}, \bibinfo{person}{Yida Wei}, \bibinfo{person}{Wangbo Yu}, \bibinfo{person}{Munan Ning}, {and} \bibinfo{person}{Li Yuan}.} \bibinfo{year}{2023}\natexlab{}.
\newblock \showarticletitle{Repaint123: Fast and high-quality one image to 3d generation with progressive controllable 2d repainting}.
\newblock \bibinfo{journal}{\emph{arXiv preprint arXiv:2312.13271}} (\bibinfo{year}{2023}).
\newblock


\bibitem[Zhang et~al\mbox{.}(2024a)]%
        {zhang2024elite-ev-gs}
\bibfield{author}{\bibinfo{person}{Zixin Zhang}, \bibinfo{person}{Kanghao Chen}, {and} \bibinfo{person}{Lin Wang}.} \bibinfo{year}{2024}\natexlab{a}.
\newblock \showarticletitle{Elite-EvGS: Learning Event-based 3D Gaussian Splatting by Distilling Event-to-Video Priors}.
\newblock \bibinfo{journal}{\emph{arXiv preprint arXiv:2409.13392}} (\bibinfo{year}{2024}).
\newblock


\bibitem[Zhang et~al\mbox{.}(2024b)]%
        {zhang2024elite-evgs}
\bibfield{author}{\bibinfo{person}{Zixin Zhang}, \bibinfo{person}{Kanghao Chen}, {and} \bibinfo{person}{Lin Wang}.} \bibinfo{year}{2024}\natexlab{b}.
\newblock \showarticletitle{Elite-evgs: Learning event-based 3d gaussian splatting by distilling event-to-video priors}.
\newblock \bibinfo{journal}{\emph{arXiv preprint arXiv:2409.13392}} (\bibinfo{year}{2024}).
\newblock


\bibitem[Zhao et~al\mbox{.}(2023)]%
        {zhao2023event-object-detection-icra23}
\bibfield{author}{\bibinfo{person}{Chunhui Zhao}, \bibinfo{person}{Yakun Li}, {and} \bibinfo{person}{Yang Lyu}.} \bibinfo{year}{2023}\natexlab{}.
\newblock \showarticletitle{Event-based real-time moving object detection based on imu ego-motion compensation}. In \bibinfo{booktitle}{\emph{2023 IEEE International Conference on Robotics and Automation (ICRA)}}. IEEE, \bibinfo{pages}{690--696}.
\newblock


\bibitem[Zou et~al\mbox{.}(2021)]%
        {zou2021event-3dhpe-cvpr21}
\bibfield{author}{\bibinfo{person}{Shihao Zou}, \bibinfo{person}{Chuan Guo}, \bibinfo{person}{Xinxin Zuo}, \bibinfo{person}{Sen Wang}, \bibinfo{person}{Pengyu Wang}, \bibinfo{person}{Xiaoqin Hu}, \bibinfo{person}{Shoushun Chen}, \bibinfo{person}{Minglun Gong}, {and} \bibinfo{person}{Li Cheng}.} \bibinfo{year}{2021}\natexlab{}.
\newblock \showarticletitle{Eventhpe: Event-based 3d human pose and shape estimation}. In \bibinfo{booktitle}{\emph{Proceedings of the IEEE/CVF International Conference on Computer Vision}}. \bibinfo{pages}{10996--11005}.
\newblock


\end{thebibliography}

\definecolor{darkgreen}{rgb}{0.0, 0.5, 0.0}
\definecolor{darkblue}{rgb}{0.0, 0.0, 0.5}

\begin{figure*}[!ht]
    \centering
    \includegraphics[width=\linewidth,trim={0cm 0cm 0cm 0cm},clip]{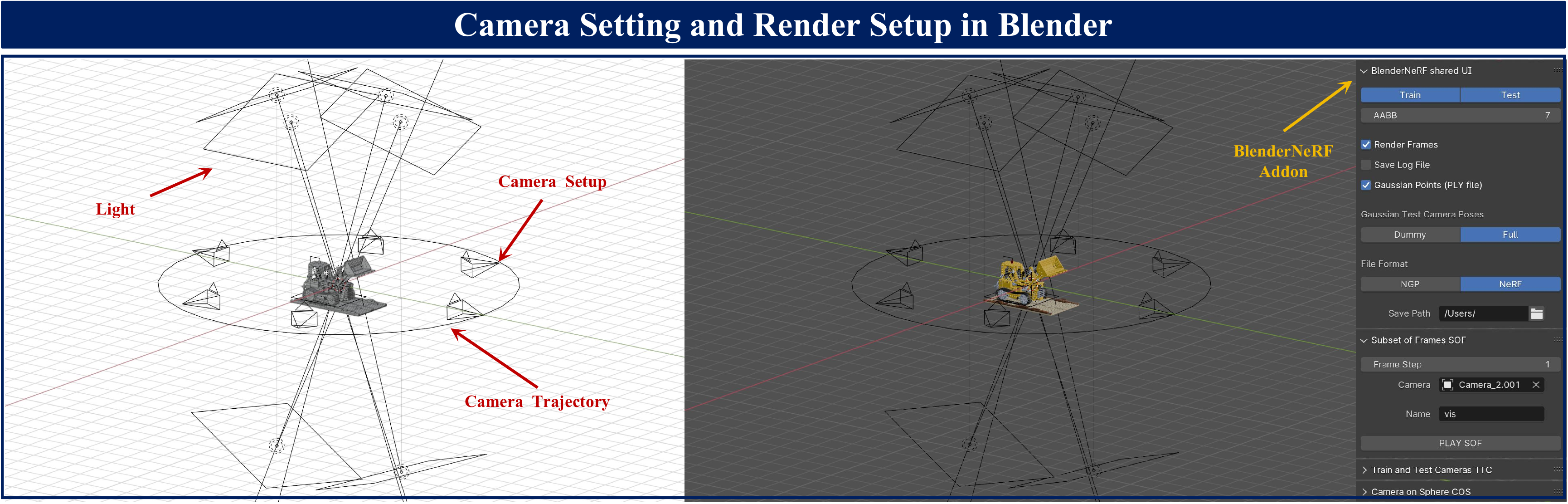}
    \vspace{-0.9em}
    \caption{
        Virtual camera setup in Blender for synthetic dataset generation. The six simulated DAVIS 346C event cameras are positioned to match the layout of our real-world multi-view recording environment.
    }
    \vspace{-0.65em}
    \label{fig:app:camera-setup}
\end{figure*}

\appendix
\label{app:data_preparation}


Detailed descriptions of dataset construction and training configurations are provided in Section~\ref{sec:app:data-preparation} of the appendix. Section~\ref{sec:app:pcl-initialization} presents the implementation of our proposed initialization strategy and compares it with existing methods. Further experimental results and ablation studies are reported in Section~\ref{sec:app:additional-experiments}. 

\section{Dataset Preparations}
\label{sec:app:data-preparation}

\begin{figure}[!h]
    \centering
    \includegraphics[width=0.9\linewidth,trim={0cm 0cm 0cm 0cm},clip]{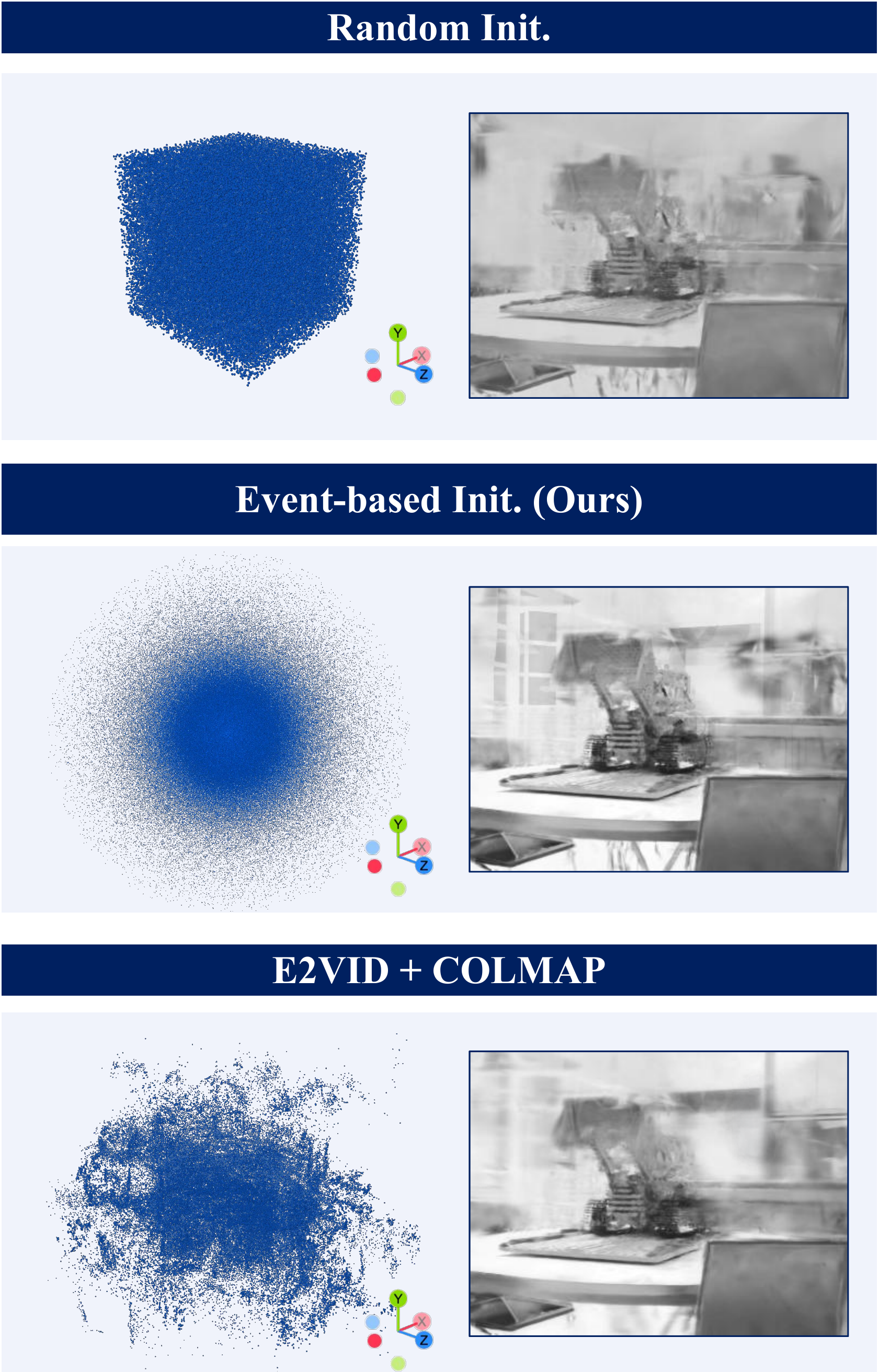}
    \vspace{-0.9em}
    \caption{
        Qualitative comparison of different initialization methods and our method achieves a trade-off between efficiency and performace.
    }
    \vspace{-0.65em}
    \label{fig:app:abltion_on_pcl}
\end{figure}

\subsection{Synthetic datasets}
\label{sec:app:data-preparation:syn}


We manually create eight synthetic scenes with six viewpoints arranged in a 360-degree configuration around the object or scene. 
Each scene is designed as a center-focus setup, with an object placed at the center. 
For these scenes, we render six dynamic scenarios at a resolution of $346 \times 260$ in Blender~\cite{Blender} at 3000 FPS with the BlenderNeRF addon~\cite{BlenderNeRF}.
Six moving viewpoints are uniformly distributed around the object in a spherical spiral motion at a constant height. 
Event streams are generated using the v2e framework~\cite{vid2e}.
Additionally, leveraging the camera trajectory data, we simulate blurry images by integrating RGB frames over the exposure time, with varying degrees of motion \textit{blur—mild}, \textit{medium}, and \textit{strong}.

For training and evaluation, we use six viewpoints for training and set the llffhold value to 8 for testing. 
For event-only dynamic reconstruction, RGB frames are converted to grayscale for evaluation, with event streams used exclusively as input. 
In the event-RGB fusion dynamic reconstruction, full-resolution color images are used in conjunction with event slices as input modalities.

\textbf{Data Composition}
The proposed synthetic dataset consists of five dynamic objects, three dynamic indoor scenes, as follows:
\begin{itemize}
    \item \textbf{Dynamic objects.} We design five object models in Blender, including \textit{Lego}, \textit{Rubik’s Cube}, \textit{MC Toy}, \textit{Hinge}, and \textit{Cubes}. The dynamic \textit{Lego} model is derived from the static \textit{Lego} in the NeRF dataset~\cite{nerf}, to which we add animation.
    \item \textbf{Dynamic indoor scenes.}  We design three indoor models with dynamic objects in Blender, including \textit{Capsule}, \textit{Restroom} and \textit{Garage}. 
\end{itemize}
All models are licensed under \textit{CC-BY 4.0} and will be open-source.

\textbf{Data Limitations.}
The synthetic data in this work is generated using the v2e framework~\cite{vid2e}, which simulates events based on images. 
However, this approach is inherently limited in handling extreme lighting conditions, such as overexposure or very low light. In these scenarios, the images themselves lack crucial information due to the nature of the lighting, which restricts the ability to accurately simulate event data for such conditions.The left is the pointcloud of Gaussian initilization and the right is the novel view of the \textit{Restroom} scene.

\begin{figure*}[!ht]
    \centering
    \includegraphics[width=\linewidth,trim={0cm 0cm 0cm 0cm},clip]{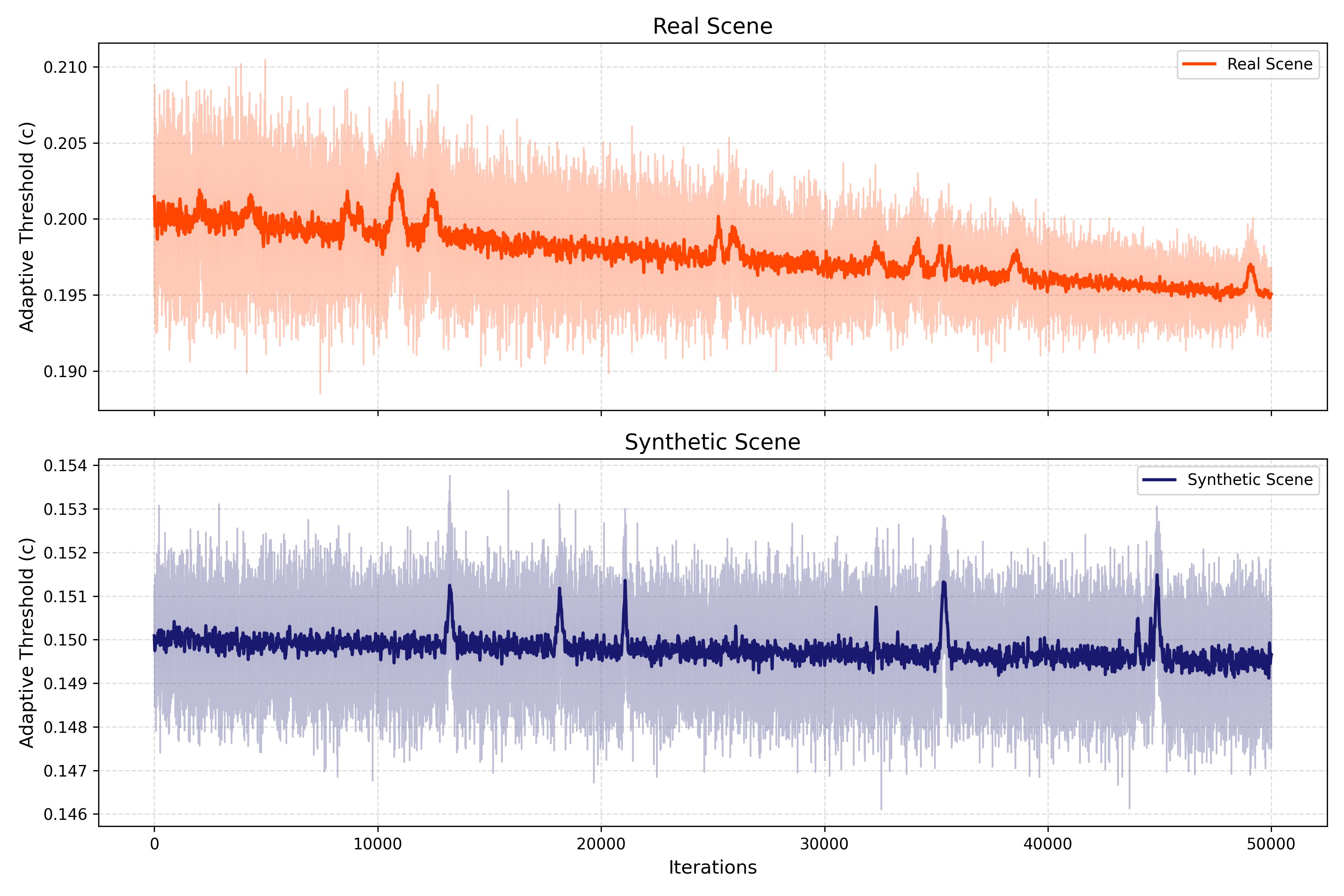}
    \vspace{-0.9em}
    \caption{
    Visualization of the adaptive event threshold $\hat{C}$ during training on both synthetic and real-scene datasets. 
    For the synthetic dataset (\textcolor{darkblue}{bottom}), $\hat{C}$ is initialized to 0.15 and remains relatively stable with occasional spikes. 
    For the real-scene dataset (\textcolor{orange}{top}), $\hat{C}$ is initialized to 0.2 and exhibits more pronounced temporal fluctuations due to sensor noise and real-world intensity transitions.
    These burst-like perturbations reflect dynamic changes in photovoltage, which are used to compute the contrast between adjacent frames. 
    A properly adjusted $\hat{C}$ is critical for robustly converting such contrast into events during training.
    }
    \vspace{-0.65em}
    \label{fig:app:abltion_on_event_threshold}
\end{figure*}

\subsection{Real-scene datasets}
\label{sec:app:data-preparation:real}

We adopt the DSEC dataset~\cite{Gehrig21ral-DSEC}, a large-scale real-world dataset designed for driving scenarios, to evaluate our method under realistic and dynamic conditions. 
The dataset was captured using a synchronized sensor rig mounted on a vehicle, consisting of a Prophesee Gen3.1 event camera, a global shutter RGB camera, and a Velodyne LiDAR. 
The event camera records asynchronous brightness changes at a spatial resolution of 640$\times$480 and provides high temporal resolution (down to microseconds), enabling the capture of fast motion and high dynamic range scenes. 
The RGB camera outputs global shutter images at 1024$\times$768 resolution with fixed frame intervals. 
Calibration files are provided to align  coordinate systems of the sensors.

Each sequence in DSEC contains temporally synchronized event streams, RGB frames, LiDAR point clouds, camera intrinsics/extrinsics, and time-stamped poses obtained via visual-inertial odometry. 
For our experiments, we select three representative sequences: \textit{interlaken\_00\_c}, \textit{interlaken\_00\_d}, and \textit{zurich\_city\_00\_a}, which cover diverse urban and suburban environments.

Since the dataset is not originally designed for Novel View Synthesis (NVS), we perform several processing steps to construct suitable input-output pairs:

\begin{itemize}
    \item \textbf{Image-Event Alignment:} For each RGB frame, we extract a corresponding event stream by accumulating events within a fixed temporal window around the image timestamp. Events outside the desired range are discarded to reduce background noise.
    \item \textbf{View Subsampling:} We uniformly sample camera viewpoints along the driving trajectory. Following the standard LLFF~\cite{mildenhall2019locallightfieldfusion-llff} protocol, we use every 8 consecutive views for training and hold out the next view for evaluation.
    \item \textbf{Modality Handling:} For event-only models, RGB frames are converted to grayscale as evaluation and only event streams are used as input. For event-RGB fusion settings, the full-resolution color images are used jointly with the event slices as input modalities.
    \item \textbf{Frame Curation:} Frames suffering from severe motion blur or under-/over-exposure are excluded to ensure a clean evaluation set. We also ignore frames with poor localization confidence based on pose metadata.
\end{itemize}

Although the dataset offers event streams, RGB images, and LiDAR data, its forward-facing setup with narrow baseline viewpoints makes it inherently unsuitable for tasks requiring diverse multi-view observations, such as high-fidelity 3D reconstruction and novel view synthesis. As a result, we use these sequences only for qualitative visualization.

\section{More details of Pointcloud Initialization}
\label{sec:app:pcl-initialization}

\begin{table}[h]
\centering
\caption{Ablation Study on Different Initialization Method.}
\label{tab:app:ablaition_init_method}
\resizebox{1.0\linewidth}{!}{
\begin{tabular}{l|ccc|c}
\toprule
\textbf{Method} & \textbf{PSNR} $\uparrow$ & \textbf{SSIM} $\uparrow$ & \textbf{LPIPS} $\downarrow$ & 
 \textbf{Time/h}\\
\midrule
Random Init. & 21.56 & 0.785 & 0.233 & 0.9 \\
E2VID+SfM & 24.87 & 0.866 & 0.170 & 2.5\\
\midrule
\rowcolor{gray!10}
\textbf{Ours} & 24.21\textsubscript{(\textcolor{red}{-0.66})} & 0.854\textsubscript{(\textcolor{red}{-0.012})} & 0.176\textsubscript{(\textcolor{red}{+0.006})} & 1.1\textsubscript{(\textcolor{darkgreen}{-1.4})}\\
\bottomrule
\end{tabular}
}
\end{table}

In this section, we explore the impact of different point cloud initialization methods on the rendering performance of \textit{E-4DGS} in three proposed indoor scenes. 
Compared to the random initialization, commonly used in methods such as ~\cite{huang2024inceventgs,wu2024ev-gs}, using the sparse point clouds from Structure-from-Motion (SfM)~\cite{micusik2006sfm} significantly improves rendering accuracy when only motion events are utilized, with the PSNR metric increasing from 24.21 dB to 24.87 dB.

To further demonstrate the trade-off between efficiency and performance achieved by our proposed method, we compare the effect of point cloud initialization using event-to-video approaches in Table ~\ref{tab:app:ablaition_init_method}.
Using E2VID\cite{Rebecq19e2vid} to convert event data into images and generating point clouds through SfM yields further accuracy improvements. 
However, this process introduces additional computational costs and time due to reliance on learning-based methods.

As shown in Figure~\ref{fig:app:abltion_on_pcl}, we visualize the impact of different initialization methods on event-based 4DGS rendering. 
When random initialization is used, the 4DGS reconstruction based on motion events suffers from noticeable artifacts and a lack of detail. 
The E2VID + COLMAP-based SfM method improves scene reconstruction, but at the cost of significantly lower runtime efficiency. 
In contrast, our method employs a radial initialization after considering a center-focus environment, yielding comparable rendering results to the two-stage initialization approach, despite slightly lower quantitative metrics. 
This validates the key role of our approach in improving the efficiency of event-driven explicit dynamic reconstruction.

\section{Additional experiments}
\label{sec:app:additional-experiments}

\subsection{Performance of adaptive event threshold}

To bridge the gap between dense image rendering and sparse event streams, our E-4DGS framework incorporates a learnable event contrast threshold $\hat{C}$.
This parameter governs the sensitivity of event triggering, and is jointly optimized with other model parameters.
Rather than relying on a fixed threshold, we allow $\hat{C}$ to dynamically evolve to better accommodate diverse temporal changes in intensity.
As shown in Fig.~\ref{fig:app:abltion_on_event_threshold}, the synthetic dataset demonstrates a relatively stable threshold behavior, aligning with its lower noise and controlled motion.
In contrast, real-scene data produces more frequent and stronger burst patterns, requiring a more adaptive threshold to handle high-frequency voltage changes effectively.
This adaptiveness ensures accurate contrast modeling for event supervision, contributing to the photometric alignment between rendered and observed event data.

\subsection{Qualitative comprisons of the motion deblurring}
\begin{figure*}[h]
    \centering
    \includegraphics[width=\linewidth,trim={0cm 0cm 0cm 0cm},clip]{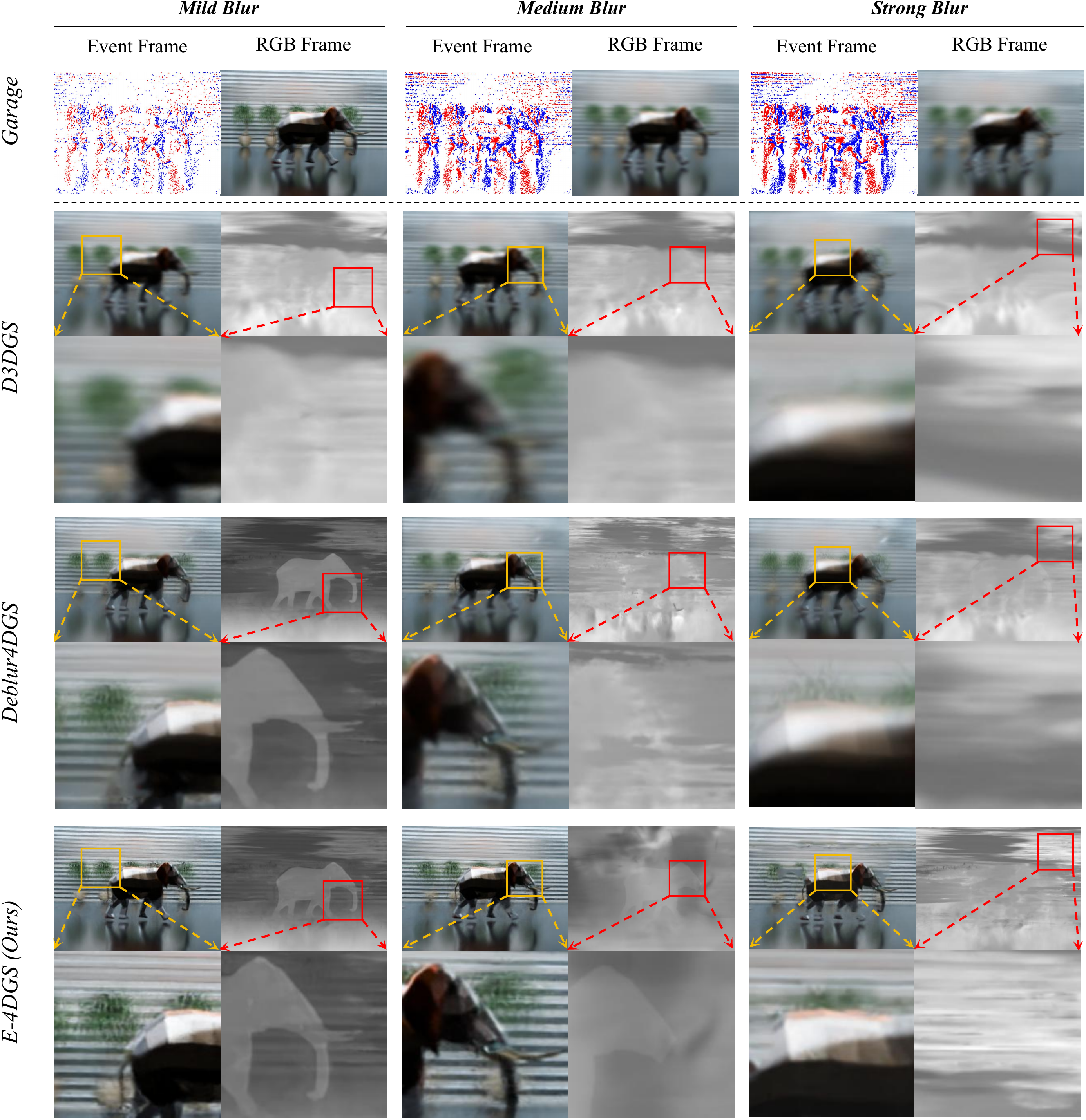}
    \vspace{-0.9em}
    \caption{
    \textbf{Qualitative comparison under varying motion blur levels on synthetic scenes.} As blur severity increases, baseline methods (\textit{D3DGS} and \textit{Deblur4DGS}) suffer from degraded reconstructions with noticeable artifacts or loss of geometric consistency. In contrast, our method (\textbf{E-4DGS}) produces high-fidelity renderings with sharper details and improved temporal coherence across all blur levels, demonstrating its robustness and effectiveness under fast motion.
    }
    \vspace{-0.65em}
    \label{fig:app:comprisons-deblur-vis}
\end{figure*}

In the main paper, we have already presented qualitative results under varying levels of motion blur. In this section, we further provide additional visual comparisons on the synthetic dataset to evaluate the robustness of different methods across mild, medium, and severe blur conditions. As shown in Figure~\ref{fig:app:comprisons-deblur-vis}, increasing blur levels degrade the reconstruction quality of baseline methods to varying degrees.
Compared to D3DGS, which struggles to recover sharp structures under heavy blur, and E2VID+D3DGS, which introduces artifacts from video reconstruction, our method E-4DGS consistently produces sharper and more temporally coherent results. Although Deblur4DGS mitigates some blur-related degradation, it lacks the geometric consistency offered by our event-guided framework. Overall, \textbf{E-4DGS} achieves high-fidelity reconstructions across all blur settings, demonstrating its robustness and effectiveness under challenging motion scenarios.

\begin{table*}[h]
\centering
\caption{Per-synthetic scene breakdown under the default setting.}
\begin{tabular}{llccccccccc}
\toprule
\multicolumn{1}{c}{} & \multicolumn{1}{c}{} & \multicolumn{8}{c}{Synthetic Scene} &  \\ \cmidrule(lr){3-10}
\multicolumn{1}{c}{\multirow{-2}{*}{Metric}} & \multicolumn{1}{c}{\multirow{-2}{*}{Method}} & \texttt{Lego} & \texttt{Rubik’s Cube} & \texttt{MC-Toy} & \texttt{Hinge} & \texttt{Cubes} & \texttt{Capsule} & \texttt{Restroom} & \texttt{Garage} & \multirow{-2}{*}{Average} \\ \midrule
 & D3DGS$_{w/o~blur}$  & 26.47 & 20.30 & 31.23 & 28.05 & 21.75 & 21.64 & 20.67 & 20.36 & 23.81 \\
 & D3DGS$_{w/~blur}$ & 23.62 & 18.12 & 27.51 & 26.46 & 19.67 & 20.08 & 19.32 & 19.05 & 21.73 \\
 & E2VID+D3DGS & 20.57 & 16.16 & 26.06 & 24.87 & 17.98 & 18.49 & 17.17 & 17.79 & 19.88 \\
 & Deblur4DGS & 23.17 & 17.68 & 28.06 & 26.35 & 19.27 & 20.10 & 19.39 & 19.23 & 21.66 \\
 & E-4DGS$_\textit{\textcolor{gray}{event}-only}$ & 26.85 & 20.97 & 31.85 & 28.83 & 22.36 & 24.23 & 23.17 & 24.81 & 25.38 \\
\multirow{-6}{*}{PSNR $\uparrow$} & \cellcolor[HTML]{F3F3F3}E-4DGS$_\textit{\textcolor{gray}{event}\& \textcolor{red}{R}\textcolor{green}{G}\textcolor{blue}{B}}$ & \cellcolor[HTML]{F3F3F3}\textbf{27.23} & \cellcolor[HTML]{F3F3F3}\textbf{21.23} & \cellcolor[HTML]{F3F3F3}\textbf{32.41} & \cellcolor[HTML]{F3F3F3}\textbf{29.02} & \cellcolor[HTML]{F3F3F3}\textbf{22.42} & \cellcolor[HTML]{F3F3F3}\textbf{24.39 }& \cellcolor[HTML]{F3F3F3}\textbf{23.30} & \cellcolor[HTML]{F3F3F3}\textbf{24.95} & \cellcolor[HTML]{F3F3F3}\textbf{25.62 }\\ \midrule
 & D3DGS$_{w/o~blur}$  & 0.099 & 0.207 & 0.077 & 0.074 & 0.129 & 0.271 & 0.278 & 0.251 & 0.173 \\
 & D3DGS$_{w/~blur}$ & 0.250 & 0.351 & 0.181 & 0.160 & 0.175 & 0.436 & 0.406 & 0.409 & 0.296 \\
 & E2VID+D3DGS & 0.346 & 0.404 & 0.267 & 0.247 & 0.298 & 0.595 & 0.527 & 0.493 & 0.397 \\
 & Deblur4DGS & 0.265 & 0.375 & 0.176 & 0.162 & 0.181 & 0.402 & 0.386 & 0.385 & 0.291  \\
 & E-4DGS$_\textit{\textcolor{gray}{event}-only}$ & 0.084 & 0.185 & 0.071 & 0.069 & 0.120 & 0.183 & 0.189 & 0.172 & 0.134 \\
\multirow{-6}{*}{LPIPS $\downarrow$} & \cellcolor[HTML]{F3F3F3}E-4DGS$_\textit{\textcolor{gray}{event}\& \textcolor{red}{R}\textcolor{green}{G}\textcolor{blue}{B}}$ & \cellcolor[HTML]{F3F3F3}\textbf{0.078} & \cellcolor[HTML]{F3F3F3}\textbf{0.172} & \cellcolor[HTML]{F3F3F3}\textbf{0.068} & \cellcolor[HTML]{F3F3F3}\textbf{0.067} & \cellcolor[HTML]{F3F3F3}\textbf{0.119} & \cellcolor[HTML]{F3F3F3}\textbf{0.178} & \cellcolor[HTML]{F3F3F3}\textbf{0.184} & \cellcolor[HTML]{F3F3F3}\textbf{0.165} & \cellcolor[HTML]{F3F3F3}\textbf{0.129} \\ \midrule
 & D3DGS$_{w/o~blur}$  & 0.910 & 0.868 & 0.956 & 0.936 & 0.924 & 0.770 & 0.765 & 0.757 & 0.861 \\
 & D3DGS$_{w/~blur}$ & 0.821 & 0.804 & 0.905 & 0.908 & 0.905 & 0.730 & 0.686 & 0.620 & 0.797 \\
 & E2VID+D3DGS & 0.765 & 0.752 & 0.851 & 0.856 & 0.852 & 0.655 & 0.547 & 0.549 & 0.728 \\
 & Deblur4DGS & 0.813 & 0.786 & 0.908 & 0.900 & 0.898 & 0.736 & 0.695 & 0.643 & 0.797 \\
 & E-4DGS$_\textit{\textcolor{gray}{event}-only}$ & 0.912 & 0.882 & 0.959 & 0.942 & 0.931 & 0.842 & 0.829 & 0.874 & 0.896 \\
\multirow{-6}{*}{SSIM $\uparrow$} & \cellcolor[HTML]{F3F3F3}E-4DGS$_\textit{\textcolor{gray}{event}\& \textcolor{red}{R}\textcolor{green}{G}\textcolor{blue}{B}}$ & \cellcolor[HTML]{F3F3F3}\textbf{0.925} & \cellcolor[HTML]{F3F3F3}\textbf{0.895} & \cellcolor[HTML]{F3F3F3}\textbf{0.963} & \cellcolor[HTML]{F3F3F3}\textbf{0.949} & \cellcolor[HTML]{F3F3F3}\textbf{0.933} & \cellcolor[HTML]{F3F3F3}\textbf{0.848} & \cellcolor[HTML]{F3F3F3}\textbf{0.835} & \cellcolor[HTML]{F3F3F3}\textbf{0.879} & \cellcolor[HTML]{F3F3F3}\textbf{0.903} \\ \bottomrule
\end{tabular}
\label{tab:breakdown}
\end{table*}

\subsection{Per-Scene Breakdown}
Table~\ref{tab:breakdown} presents the quantitative results of all methods for each of the eight synthetic scene sequences, simulated with default settings that are optimal for all methods. The per-scene results are generally consistent with the aggregate metrics, as discussed in Section 5.1.2. Our method outperforms the baselines in most scenes and shows comparable performance in others.

\section{Broader Impact and Limitations}
\label{sec:app:Impact-limitation}
\noindent \textbf{Broader Impact.} 
The proposed \textit{E-4DGS} framework opens up new possibilities for high-fidelity 4D reconstruction in domains where traditional cameras fall short due to motion blur or limited dynamic range. 
By leveraging the high temporal resolution of event cameras, our method enables temporally coherent scene modeling under rapid motion, which is beneficial for a variety of real-world applications including autonomous robotics, high-speed inspection, sports analytics, and scientific visualization in challenging illumination conditions.
Furthermore, the ability to reconstruct dynamic scenes using purely event-based supervision contributes to the development of low-latency, power-efficient visual systems, which are particularly relevant for resource-constrained or edge computing scenarios.

\noindent \textbf{Limitations.}
While \textit{E-4DGS} demonstrates promising results in dynamic 3D scene reconstruction, certain scenarios, such as those involving extreme motion or significant occlusions, may present challenges for the method. The performance is highly dependent on the availability of synchronized multi-view event data and precise camera calibration. These aspects are areas for further exploration to enhance robustness and generalizability in more complex environments.

\noindent \textbf{Project Release.} 
We implemented \textit{E-4DGS} based on the official code of Deformable3DGS~\cite{yang2024deformable3dgs}, Gaussianflow~\cite{gaussianflow}, E-NeRF~\cite{klenk2023e-nerf} and Event3DGS~\cite{han2024event3dgs-nips} with Pytorch
Upon the publication of the paper, we will release the project  materials.


\end{document}